\documentclass{article}

  \usepackage[preprint]{neurips_2026}

\usepackage[utf8]{inputenc} 
\usepackage[T1]{fontenc}    
\usepackage{hyperref}       
\usepackage{url}            
\usepackage{booktabs}       
\usepackage{amsfonts}       
\usepackage{nicefrac}       
\usepackage{microtype}      
\usepackage{xcolor}         

\usepackage{makecell}
\usepackage{threeparttable}
\usepackage{graphicx}

\usepackage{adjustbox} %
\usepackage{booktabs}  %

\usepackage{amsmath,amsfonts}
\usepackage{multirow}

\usepackage{algorithm}
\usepackage{algorithmicx}
\usepackage{algpseudocode}

\newcommand{\bxhat}{\hat{\mathbf{x}}}

\newcommand{\cDP}{\mathcal{C}_{\mathrm{DP}}}
\newcommand{\cD}{\mathcal{C}_{\mathrm{D}}}
\newcommand{\cP}{\mathcal{C}_{\mathrm{P}}}
\newcommand{\JRDP}{J_{\mathrm{RDP}}}
\newcommand{\gc}{g_c}
\newcommand{\gth}{g_\theta}
\newcommand{\epsth}{\epsilon_\theta}
\newcommand{\ROICRD}{\text{DCIC}_{\text{RD}}}
\newcommand{\ROICRP}{\text{DCIC}_{\text{RP}}}
\newcommand{\ROICRDP}{\text{DCIC}_{\text{RDP}}}

\title{Dual-Constrained Diffusion Image Compression for Operational Rate-Distortion-Perception Optimization}

\author{%
  Sanxin Jiang \\
  Department of Information Engineering\\
  Shanghai University of Electric Power\\
  Shanghai, China \\
  \texttt{samjoe\_2018@shiep.edu.cn} \\
   \And
   Jiro Katto \\
   Department of Computer Science \\ 
   and Communication Engineering \\
    Waseda University \\
   Tokyo, Japan \\
   \texttt{katto@waseda.jp} \\
   \AND
   Heming Sun \\
   Faculty of Engineering \\
    Institute of Science Tokyo \\
   Tokyo, Japan \\
   \texttt{son.k.2b4a@m.isct.ac.jp} \\
}

\begin{document}

\maketitle

\begin{abstract}
The rate--distortion--perception (RDP) trade-off extends classical rate--distortion theory by imposing a distributional constraint on reconstructions, providing a unified framework for neural image compression that jointly governs fidelity and perceptual realism. While prior work has achieved near-optimal rate--perception trade-offs, practical approaches that explicitly realize the full RDP trade-off remain scarce, primarily due to the difficulty of introducing common randomness at the decoder. We propose \textbf{DCIC} (\textbf{D}ual-\textbf{C}onstrained {D}iffusion \textbf{I}mage \textbf{C}ompression), a framework that integrates a learned image compression codec with a diffusion-based decoder governed by joint distortion and idempotence constraints. The distortion constraint bounds reconstruction fidelity relative to the base codec output, while the idempotence constraint — requiring that re-encoding the restored image recovers the base codec reconstruction — serves as a tractable surrogate for the distributional perception requirement; together, they guide the reverse denoising process via iterative optimization with consistent noise injection, realizing common randomness without additional rate overhead. At a fixed rate, the dual constraints jointly navigate the Pareto frontier of the distortion--perception (D,P) plane via attenuation factors $(K_D, K_P)$, enabling multiple reconstructions of continuously adjustable fidelity--realism from a single bitstream. DCIC$_{\text{RD}}$($K_P\!=\!0$) and DCIC$_{\text{RP}}$($K_D\!=\!0$) are subsumed as boundary curves on this frontier, with DCIC$_{\text{RDP}}$($K_D\!=\!K_P\!=\!1$) realizing the optimal interior operating point. 
Extensive experiments on CelebA-HQ, CLIC2020, and ImageNet-1K across CNN, Transformer, and hybrid architectures demonstrate that DCIC$_{\text{RDP}}$ achieves superior BD-PSNR over all perceptual codecs while DCIC$_{\text{RP}}$ matches dedicated perception-oriented methods in BD-FID, confirming the practical value of full RDP surface navigation.
\end{abstract}

\section{Introduction}

Classical rate--distortion (RD) theory frames image compression as the
problem of reconstructing a source signal that closely approximates the
original while satisfying a rate constraint~\cite{sullivan2012hevc}.
End-to-end learned image compression (LIC) methods~\cite{cheng2020learned,liu2023learned}
have made notable advances in RD performance, with state-of-the-art
approaches~\cite{he2022elic,wang2024variable,Sun2025QLIC}
surpassing handcrafted codecs such as VVC~\cite{bross2021vvc} in
terms of PSNR and MS-SSIM.

However, MSE-based distortion metrics fail to capture perceptual
quality~\cite{blau2019rethinking,chen2025information}.
Blau and Michaeli~\cite{blau2019rethinking} introduced the information
rate--distortion--perception (RDP) function, formalizing the three-way
trade-off among coding rate, distortion, and perceptual quality by
imposing a distributional constraint on reconstructions.
Within this framework, perfect perception corresponds to the case
where the reconstruction distribution exactly matches the source distribution.

Meeting the realism constraint generally necessitates stochastic
decoding~\cite{wagner2022rdp}.
Theoretical analysis~\cite{chen2022rdp,qian2025rdpgaussian}
predicts that shared high-quality common randomness between encoder
and decoder could benefit lossy compression, yet this has not been
observed in practical systems.
Most state-of-the-art perceptual image codecs
(PIC)~\cite{wagner2022rdp,liu2024rdcognition,chen2024survey}
inject independent noise at the decoder, which unavoidably increases
distortion~\cite{qian2025rdpgaussian,wang2025taskoriented},
underscoring the inherent fidelity--realism tension.

GAN-based approaches~\cite{mentzer2020hific,agustsson2023multirealism}
improve perceptual quality by augmenting RD objectives with adversarial
and LPIPS terms.
Diffusion-based codecs~\cite{yang2023cdc,xu2024idempotence,jiang2025rddm,xu2025picd}
offer further gains at low bitrates, but typically do not provide a
comprehensive characterization of the full RDP surface nor explicit
common randomness.


In this paper we address both challenges simultaneously. Our contributions are:
\begin{itemize}
	\item \textbf{Operational RDP formulation.} We formulate the RDP trade-off as a distortion--perception constrained bi-objective optimization problem at a fixed rate, deriving an explicit per-step objective grounded in the conditional perception measure~\cite{salehkalaibar2024rdp,niu2023conditional}. This is the first practical framework formally connected to the operational $R_C(D,P)$ function.
	
	\item \textbf{Joint distortion--idempotence constraints.} We show that jointly imposing a distortion constraint $\cD$ — bounding MSE between the restored and base-codec reconstruction — and an idempotence constraint $\cP$ — requiring that re-encoding the restored image recovers the base-codec output, thereby satisfying the distributional perception requirement — on the diffusion reverse process is both necessary and sufficient to navigate the Pareto frontier of the (D,P) plane. Neither constraint alone achieves this; their joint formulation is the theoretical core of this work.
	
	\item \textbf{Common randomness without rate overhead.} Consistent noise injection via the base codec $g_c$ across encoding and decoding realizes shared randomness within the diffusion reverse process, satisfying the Markov chain requirement of the RDP function without transmitting additional bits.
	
	\item \textbf{Hierarchical fidelity--realism control.} Attenuation factors $(K_D,\,K_P)\in[0,1]^2$ enable continuous navigation of the (D,P) Pareto frontier from a single bitstream. DCIC$_{\text{RD}}$($K_P\!=\!0$) and DCIC$_{\text{RP}}$($K_D\!=\!0$) are subsumed as boundary curves, with DCIC$_{\text{RDP}}$($K_D\!=\!K_P\!=\!1$) realizing the optimal interior operating point.
	
	\item \textbf{State-of-the-art performance and generalizability.} Across CNN, Transformer, and hybrid LIC codecs on CelebA-HQ, CLIC2020, and ImageNet-1K, DCIC$_{\text{RDP}}$ achieves superior BD-PSNR over all perceptual codecs and DCIC$_{\text{RP}}$ matches dedicated perception-oriented methods in BD-FID, with strong architectural generalizability.
\end{itemize}

%
%
%

\section{Related Works}

\subsection{Learned Image Compression}
End-to-end LIC builds on variational autoencoders with entropy coding,
originating from the scale hyperprior framework~\cite{balle2018variational}.
Subsequent advances introduced discretized Gaussian mixture
likelihoods~\cite{cheng2020learned},
channel-conditional entropy models~\cite{he2022elic},
Transformer-based entropy coding~\cite{qian2022entroformer},
and hybrid CNN--Transformer designs~\cite{liu2023learned,wu2020hybrid}.
These methods optimize the RD trade-off but do not explicitly control
perceptual quality.

\subsection{Perceptual Image Compression}
HiFiC~\cite{mentzer2020hific} augmented LIC with a conditional GAN.
Agustsson et al.~\cite{agustsson2023multirealism} extended ELIC with
adversarial and LPIPS terms.
ILLM~\cite{muckley2023illm} incorporated implicit local likelihood
models for statistical fidelity.
These methods improve RP trade-offs but do not address the full RDP surface.

\subsection{Diffusion-Based Compression}
CDC~\cite{yang2023cdc} employed diffusion conditioned on quantized
latents.
IPIC~\cite{xu2024idempotence} and RDDM~\cite{jiang2025rddm}
introduced idempotence-based refinement of LIC outputs without training
new diffusion models.
These approaches yield visually compelling results but lack a formal
characterization of the operational RDP function or explicit control of common
randomness.
Our work builds on these foundations while providing a theoretically
grounded and comprehensive RDP framework.

\section{Problem Formulation}
\label{sec:formulation}
\subsection{Operational RDP Function and Distortion-Perception Constraints}
Let $x\sim p_x$ be the source.
A stochastic encoder $f\colon x^n\to\mathcal{M}$ and decoder
$g\colon\mathcal{M}\to\tilde{x}^n$ define a lossy codec.
A rate $R$ is \emph{achievable} under distortion constraint $D$ and
perception constraint $P$ if:
\begin{equation}
	R_C(D,P): 
	\begin{cases}
		\tfrac{1}{n}\mathbb{E}[\ell(M)]\le R,\quad \\
		\tfrac{1}{n}\mathbb{E}[\Delta(x^n,\tilde{x}^n)]\le D,\quad \\
		\tfrac{1}{n}\mathbb{E}[\varphi(p_{x^n|M},p_{\tilde{x}^n|M})]\le P.
	\end{cases}
\end{equation}
Here, $\ell(M)$ denotes the codeword length of $M$.  
The infimum of all such rates $R$ is denoted by $R_{C}(D, P)$, 
referred to as the \emph{operational} $R\!D\!P$ function corresponding to 
the conditional-distribution-based perception measure. 

Assume $| \mathcal{X} | < \infty$. For $D \geq 0$ and $P \geq 0$, $R_{C}(D, P)$ 
equals to  the following informational RDP function \cite{salehkalaibar2024rdp}:
\begin{align}
	R_C(D,P) = \inf_{p_{\hat{x},\tilde{x}|x}} I(x;\hat{x}) \quad\quad
	\text{s.t.}~~ %
	& x\leftrightarrow\hat{x}\leftrightarrow\tilde{x} \text { form a Markov chain},\; \notag \\
	& \mathbb{E}[\Delta(x,\tilde{x})]\le D,\; 
	\quad\mathbb{E}[\varphi(p_{x|\hat{x}},p_{\tilde{x}|\hat{x}})]\le P. \label{eq:rdp} 
\end{align}
The auxiliary random variable $\hat {x}$ serves as a representation of ${x}$, 
directly corresponding to the encoder output $M$ in the operational formulation.
Furthermore, function $R_C(D, P)$ is convex in $(D, P)$.

Given a base codec $\gc$ yielding reconstruction $\hat{x}$ from source
$x$, a generative refinement model $\gth$ produces restored image
$\tilde{x}$.
For an ideal restoration, it must satisfy at least two requirements. 
First, the distortion between the restored image and the reconstruction 
should be no greater than the MSE of the codec. 
Second, if the restored image is subsequently re-encoded using the base codec, 
the resulting reconstruction should be identical to the original reconstruction  $\hat{x}$.
Consequently, the combined distortion--perception constraint $\cDP$ is:
\begin{equation}
	\cDP:\;\begin{cases}
		\cD:\;\mathbb{E}[\Delta(\tilde{x},\hat{x})]\le D^* & \text{(distortion)}\\
		\cP:\;\gc(\tilde{x})=\hat{x}                        & \text{(idempotence)}
	\end{cases}
	\label{eq:cdp}
\end{equation}
$\cD$ bounds the MSE between restored and base-codec reconstruction;
$\cP$ enforces idempotence — re-encoding $\tilde{x}$ through $\gc$
recovers $\hat{x}$.
As idempotence tightens, the conditional distributions
$p_{x|\hat{x}}(\cdot|\hat{x})$ and $p_{\tilde{x}|\hat{x}}(\cdot|\hat{x})$
converge, satisfying the perception constraint in Eq.~\eqref{eq:rdp}.

\subsection{Distortion–Perception Objective via Diffusion-Guided Optimization}
In $T$ time steps, diffusion model~\cite{dhariwal2021diffusion} transforms data $x_0\sim p_x$ 
into Gaussian noise $x_T\sim\mathcal{N}(0,I)$ by iteratively adding
noise, then reverses via a learned denoiser $\epsth$.
The reverse process is a Markov chain:
\begin{equation}
	p_\theta(x_{t-1}|x_t)=\mathcal{N}(x_{t-1};\mu_\theta(x_t,t),\Sigma_\theta I).
\end{equation}
Here, the mean, 
represented by ${\mu}_{\theta}\left({x}_{t}, t\right)$, 
is the target we aim to estimate using a neural network, 
denoted by ${\epsilon}_{\theta}$, and the variance, 
denoted by $\Sigma_{\theta} $, 
can be either time-dependent constants or learnable parameters.
%
Under the DDIM one-step approximation~\cite{song2020ddim}, the
clean-sample prediction is:
\begin{equation}
	\tilde{x}_0 = f_\theta(x_t)
	= \frac{x_t - \sqrt{1-\bar\alpha_t}\,\epsth(x_t,t)}{\sqrt{\bar\alpha_t}},
	\label{eq:ddim}
\end{equation}
which we exploit for differentiable constraint evaluation during
iterative optimization.

As the RDP trade-off is inherently associated with randomness \cite{{blau2019rethinking}}, 
we employ a diffusion model to introduce the required stochasticity. 
Specifically, during its reverse process over $T$ time steps, 
the diffusion model begins with a Gaussian noise sample 
$\tilde{{x}}_T \sim \mathcal{N}(\mathbf{0}, \mathbf{I})$ and iteratively denoises it, 
ultimately producing a clean sample $\tilde{{x}}_0$ (i.e., the recovered image $\tilde{{x}}$), 
accompanied by a sequence of intermediate 
variables $\tilde{{x}}_{T-1}, \tilde{{x}}_{T-2}, \dots, \tilde{{x}}_1$.  
Throughout this generative process, we require that all elements of the generated sequence simultaneously satisfy the two constraints specified in Equation~\eqref{eq:cdp}, with the highest possible probability.  
Accordingly, 
to maximize the probability of the generated sequence satisfying
$\cDP$, we define:
\begin{equation}
	J'_{\mathrm{DP}}=
	\max_{\tilde{x}_T\sim\mathcal{N}(0,I)}
	p_\theta(\tilde{x}_T,\tilde{x}_{T-1},\ldots,\tilde{x}_0\mid\cDP).
	\label{eq:jrdp_raw}
\end{equation}
Here, $\mathcal{J}^{\prime}_{\text{DP}}$ denotes the distortion–perception objective function, 
$\theta$ represents the parameters of the diffusion model.

Equation~\eqref{eq:jrdp_raw} can be interpreted as a decoder 
matching the deterministic encoder of the basic-codec.  
Unlike a conventional decoder, it introduces randomness during the reconstruction process and enforces both distortion and perceptual constraints, thereby implementing an $R(D, P)$ function.  
Since the image $\hat{{x}}$ is reconstructed by the base codec and depends solely on the source image ${{x}}$, 
while the image $\tilde{{x}}$ is generated by the diffusion model and depends only on $\hat{{x}}$, 
the triplet $({{x}}, \hat{{x}}, \tilde{{x}})$ forms a Markov chain 
that satisfies the constraints specified in Equation~\eqref{eq:rdp}.  
It is important to note that the introduced randomness is applied consistently across the entire base codec, 
yielding a form of shared randomness.  
Consequently, Equation~\eqref{eq:jrdp_raw} effectively corresponds to the $R_C(D, P)$ function,
which achieves the optimal RDP trade-off.

Taking the negative log and exploiting the Markov property gives the
per-step objective (derivation in supplementary Sec.~A):
\begin{equation}
	J^{(t)}_{\mathrm{DP}}\approx
	\frac{1}{2\xi_t^2}\!\left[
	\|\bxhat-\gc(\tilde{x}_0)\|^2 + \|\bxhat-\tilde{x}_0\|^2
	\right]
	+\frac{1}{2\sigma_t^2}\|\tilde{x}_t-\mu_t\|^2+K. 
	\label{eq:jt}
\end{equation}
The gradient w.r.t.\ $\tilde{x}_t$, using $\partial\gc/\partial\tilde{x}_0\approx 1$
(valid at sufficiently high bitrates), is:
\begin{equation}
	\nabla_{\tilde{x}_t}J^{(t)}_{\mathrm{DP}}\approx
	\frac{2}{\xi_t^2}\cdot\frac{\partial f_\theta(\tilde{x}_t)}{\partial\tilde{x}_t}
	\cdot\!\left[\bxhat-\frac{\gc(\tilde{x}_0)+\tilde{x}_0}{2}\right]
	+\frac{1}{\sigma_t^2}(\tilde{x}_t-\mu_t).
	\label{eq:grad}
\end{equation}
This gradient has two components: (i) a joint distortion--perception
term coupling $\tilde{x}_0$ to both $\gc$ and $\hat{x}$, which
propagates common randomness from the encoder into decoding, and
(ii) a denoising prior term.

\section{DCIC Decoder}
\label{sec:decoder}

\subsection{Bi-Objective Optimization  and Decoding Architecture}

\begin{figure*}[!th]
	\centering
	\includegraphics[width=5.42in]{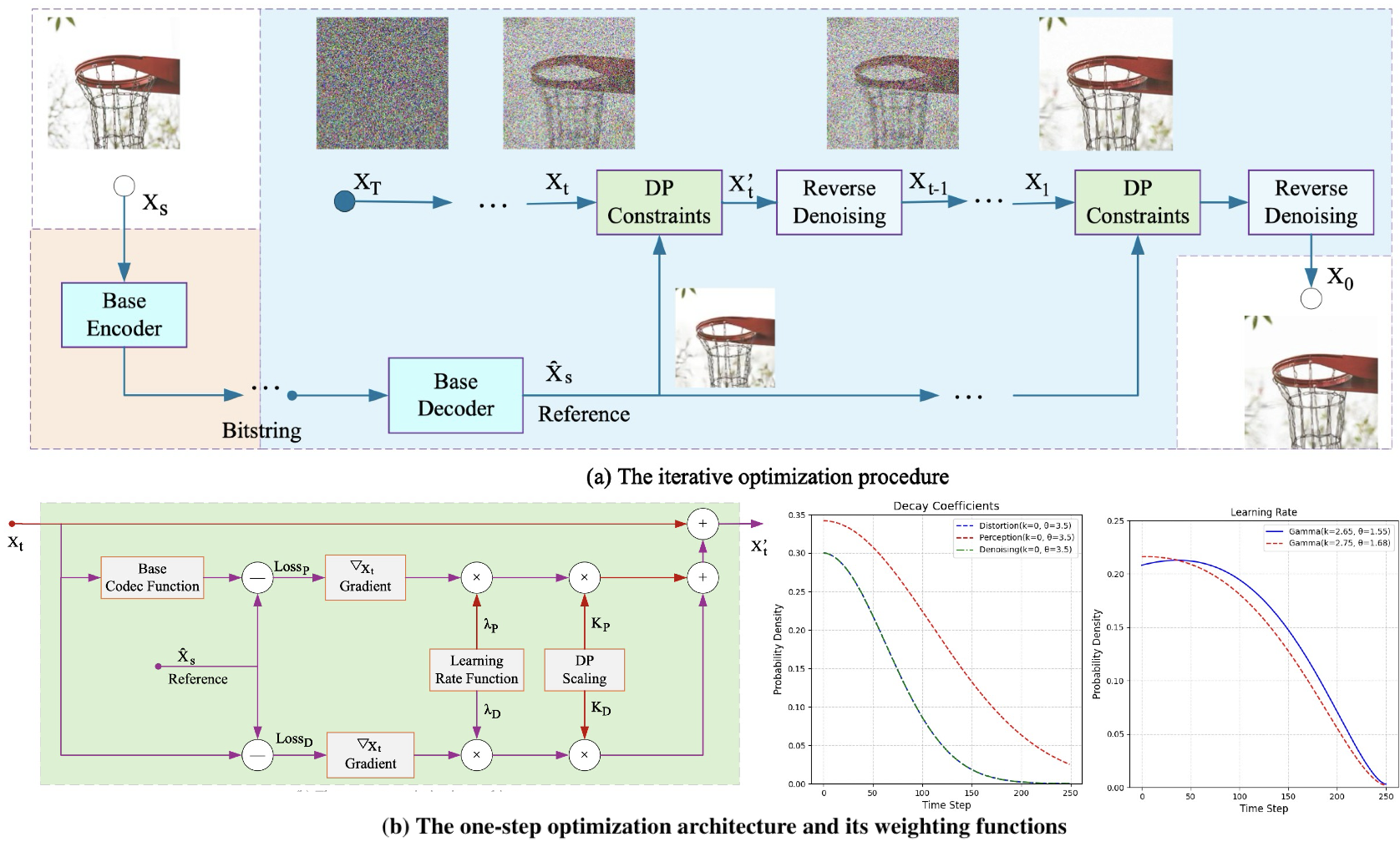}
	\caption{Overview of the DCIC architecture. The reconstruction $\hat{\boldsymbol{x}}_{0}$ from a base codec, together with the distortion–perception constraints $\mathcal{C}_{\text{DP}}$, guides the optimization of each step in the diffusion model’s reverse denoising process. Subfigures (a) and (b) illustrate the iterative optimization procedure and the one-step optimization architecture of DCIC, respectively.
	}
	\label{fig:framework}
\end{figure*}

To obtain the optimal restored image, 
we employ gradient descent to minimize the target Equation~\eqref{eq:jt}. 
Specifically, during the iterative process from time step $T$ to time step $0$, 
the gradient is progressively reduced, approaching zero at the final step. 
A learning rate function $\eta(t)$ is introduced to regulate the gradient magnitude at each time step, 
thereby ensuring a controlled and sequential decrease throughout the optimization process. 
Furthermore, to dynamically regulate the gradient variation at each time step and thereby achieve an optimal compromise 
between fidelity and realism, 
we introduce two weighting functions, $\lambda_{\text{D}}(t)$ and $\lambda_{\text{P}}(t)$, 
into the objective function $\mathcal{J}_{\text{DP}}^{(t)}$. 
These functions are applied to the distortion constraint and the perceptual constraint, respectively. 
Accordingly, the gradient of the objective function can be expressed as:
%
\begin{equation}
	\nabla_{\tilde{x}_t}J^{(t)}_{\mathrm{DP}}=
	\eta(t)\!\left[
	\lambda_D(t)(\bxhat-\tilde{x}_0)+\lambda_P(t)(\bxhat-\gc(\tilde{x}_0))
	\right]
	+\lambda_M(t)(\tilde{x}_t-\mu_t).
	\label{eq:weighted_grad}
\end{equation}
Here,  $\lambda_\text{M}(t)$ represent the decay coefficient of denoising at time  step $t$. 
Then, the architecture of DCIC is presented in Fig.~\ref{fig:framework}.


As shown in Equation~\eqref{eq:weighted_grad}, 
the two constraints converge toward different gradient endpoints: the perception term $\left[\hat{{x}}-{{g}_c\left(\tilde{{x}}_{0}\right)}\right]$ must approach zero to satisfy $\mathcal{C}_{\text{P}}$, 
whereas the distortion term $\left[\hat{{x}}-\tilde{{x}}_{0}\right]$ need 
only remain within $2 \Delta^*$ to satisfy $\mathcal{C}_{\text{D}}$. 
Consequently, $\lambda_{\text{P}}(t)$ must decay more rapidly than $\lambda_{\text{D}}(t)$.
As a result, the perception weighting curve in Fig.~\ref{fig:framework}(b) lies consistently above the distortion curve.

\subsection{Learning Rate and Weighting Functions}

To ensure stable optimization, the learning rate must increase as the
time step decreases. Near step $T$, the low SNR of $\mathbf{x}_t$
produces large gradient magnitudes, necessitating a smaller learning
rate for stability; near step $0$, the high SNR yields small gradients,
where a larger learning rate accelerates convergence toward zero.
To accommodate these opposing dynamics, we design a learning rate
schedule based on a segment of the gamma distribution, and employ a
segment of the normal distribution as the time-dependent weighting
functions for the distortion and perception constraints:
\begin{equation}
	\eta(x;\,k,\theta)
	=\frac{x^{k-1}e^{-x/\theta}}{\theta^{k}\,\Gamma(k)},
	\qquad
	\lambda(x;\,k,\sigma)
	=\frac{k}{\sigma\sqrt{2\pi}}
	\exp\!\left(-\frac{x^{2}}{2\sigma^{2}}\right),
	\label{eq:schedules}
\end{equation}
where $\Gamma(k)=\int_{0}^{\infty}t^{k-1}e^{-t}\,\mathrm{d}t$ is the
Gamma function, $k$ and $\theta$ are the shape and scale parameters of
$\eta$, and $\sigma$ controls the shape of $\lambda$ while $k$
determines its scale.
Both functions share the desired property of being small at early
time steps and increasing toward later ones, providing smooth and
consistent constraint modulation throughout the reverse process.
In practice, $\eta$ is sampled over $[0,3]$ at 250 evenly spaced
points, and $\lambda$ is sampled over $[0,8]$ at 250 evenly spaced
points, yielding per-step coefficients for each time step.
Representative curves are illustrated in Fig.~\ref{fig:framework}(b).

\subsection{Hierarchical Control via Attenuation Factors}
\label{sec:attenuation}
Hierarchical fidelity--realism control is achieved by attenuating the relative strengths of the two constraints via scalar factors  $K_D,K_P\in[0,1]$. Let $\lambda^{\mathrm{OPT}}_D(t)$
and $\lambda^{\mathrm{OPT}}_P(t)$ denote the distortion and perceptual weights 
at the optimal RDP operating point of $\mathcal{J}_{\mathrm{RDP}}$. 
The scaled weights are then defined as:
\begin{align}
	\lambda_{\text{D}}(t) = K_{\text{D}} \cdot \lambda^{\text{OPT}}_{\text{D}}(t), 
	\qquad 
	\lambda_{\text{P}}(t) = K_{\text{P}} \cdot \lambda^{\text{OPT}}_{\text{P}}(t). 
\end{align}
Although $K_{\text{D}}$ and $K_{\text{P}}$ are formally independent, 
we impose the condition that attenuation of one constraint is always accompanied 
by the other remaining at its optimal level, ensuring the restored image remains practically meaningful. 
This yields three representative cases:

\begin{itemize}
	\item \textbf{\( K_{\text{P}} = 0 \) (or $K_{\text{D}} = 0$):} 
	One constraint is completely removed, reducing the optimization to a single-constraint problem. 
	Setting $K_{\text{P}} = 0$ yields $\text{DCIC}_\text{RD}$, 
	which optimizes the RD trade-off with MSE as the sole constraint, 
	resembling a conventional codec. 
	Setting $K_{\text{D}} = 0$ 
	yields $\text{DCIC}_\text{RP}$, which optimizes the RP trade-off with idempotence as the sole constraint, 
	falling within the paradigm of perceptual image compression. 
	Both are degenerate cases of DCIC corresponding to the two extreme endpoints of the RDP surface.
	
	\item \textbf{$0 < K_{\text{P}} < 1$ (or $0 < K_{\text{D}} < 1$):} 
	Partial attenuation of one constraint while holding the other 
	at its optimal level enables graded fidelity--realism control. 
	Since the distortion constraint exerts a more dominant influence on 
	reconstruction quality than the perceptual constraint at a fixed rate, 
	$K_{\text{D}}$ is discretized more finely into $\{1/2, 1/4, 1/8, 1/16\}$, 
	while $K_{\text{P}}$ takes the coarser levels $\{1/2, 1/4\}$.
	
	
	\item \textbf{\( K_{\text{P}} = K_{\text{D}} = 1 \):} 
	Both constraints are fully active, yielding $\text{DCIC}_{\text{RDP}}$ — the optimal RDP operating point.
	
\end{itemize}

As illustrated in Fig.~\ref{fig:framework}(b), the attenuation factors 
$K_{\text{D}}$ and $K_{\text{P}}$ are applied within the DCIC decoding architecture to 
modulate the relative strengths of the two constraints. 
By varying ($K_{\text{D}}$, $K_{\text{P}}$) without retraining, 
DCIC can generate multiple reconstructions of continuously adjustable 
fidelity and perceptual quality from a single bitstream — a capability absent in existing one-to-one codecs.

\subsection{Training Protocol}
\label{sec:training}
Since the source, reconstructed, and restored images form a Markov chain (Eq.~\eqref{eq:rdp}), the restored image has no direct dependence on the source, precluding end-to-end training of the full model. Accordingly, the two network components of DCIC — the base codec ${g}_{c}$ and the diffusion model ${\epsilon}_{\theta}(x_{t}, t)$ — are trained separately.

Hyperparameter tuning in DCIC is guided by examining the two constraint extremes. Optimal fidelity requires the distortion constraint gradient to decrease within [0,$\Delta^{\star}$], while optimal realism requires the perception constraint gradient to approach zero. Since fidelity and realism are mutually coupled, the two objectives are tuned alternately — fixing one constraint while optimizing the other — until convergence.

\section{Experiments}
\label{sec:experiments}

\subsection{Setup}
\label{sec:setup}
\noindent\textbf{Datasets.}
Following~\cite{zhang2023inpainting,Kawar2022DenoisingDR}, we evaluate on three benchmarks: \textbf{CelebA-HQ}~\cite{liu2015celeba} (split from~\cite{suvorov2022lama}), \textbf{ImageNet-1K}~\cite{russakovsky2015imagenet} (original split), and \textbf{CLIC2020}~\cite{toderici2020clic}. All images are center-cropped to $256 \times 256$.

\noindent\textbf{Base codecs.} 
DCIC requires only that the base codec be differentiable, imposing no further architectural constraints. We primarily evaluate with \textit{Entroformer}~\cite{qian2022entroformer} (CNN-based) and report its full RDP trade-off surface. To demonstrate generalizability, we additionally integrate DCIC with \textit{SwinT}~\cite{zhu2022transformer} (Transformer-based) and \textit{TCM}~\cite{wu2020hybrid} (hybrid CNN--Transformer).

\noindent\textbf{Diffusion models.}
We use the diffusion model of Lugmayr et al.~\cite{lugmayr2022repaint} for CelebA-HQ, and that of Dhariwal et al.~\cite{Dhariwal2021DiffusionMB} for ImageNet-1K and CLIC2020. All pretrained models are publicly available.


\noindent\textbf{Metrics.} Fidelity is measured by PSNR and MS-SSIM; realism by LPIPS~\cite{zhang2018lpips} and FID~\cite{heusel2017fid}. For fair cross-codec comparison at varying bitrates, we report BD-PSNR, BD-LPIPS, and BD-FID~\cite{bjontegaard2001psnr,zhang2018lpips,heusel2017fid} as the primary evaluation metrics. All metrics are computed in RGB.

For all DCIC configurations, the number of reverse sampling steps is fixed at $T=250$. 
Hyperparameters are tuned on the first 50 validation images; 
final evaluation is performed on the first 500 test images.

\subsection{RDP Trade-off Surface}
\label{sec:rdp_surface}

To construct the full RDP trade-off surface, we set $K_D \in \{1, 1/2, 1/4, 1/8, 0\}$ 
and $K_P \in \{1, 1/2, 0\}$. 
Under the constraint that at least one factor remains at its optimal level, 
this yields seven valid $(K_D, K_P)$ combinations — $\{1,1\}$, $\{1,0\}$, $\{0,1\}$, 
$\{1,1/2\}$, $\{1,1/4\}$, $\{1,1/8\}$, $\{1/2,1\}$ 
— corresponding to $\text{DCIC}_\text{RDP}$, 
$\text{DCIC}_\text{RD}$, 
$\text{DCIC}_\text{RP}$, 
$\text{DCIC}_{K_D}(1/2)$, 
$\text{DCIC}_{K_D}(1/4)$, 
$\text{DCIC}_{K_D}(1/8)$, 
and $\text{DCIC}_{K_P}(1/2)$. 
The resulting RDP surface and Pareto front curves on CLIC2020 (0.1152--0.9868~bpp) are shown in Fig.~\ref{fig:pareto_surface}(a)--(b); RD and RP curves are provided in supplementary Sec.~C.

Three observations emerge from Fig.~\ref{fig:pareto_surface}. 
First, the RDP surface is convex over (D, P), 
consistent with the theoretical convexity of R(D,P)~\cite{salehkalaibar2024rdp}. 
Second, at any fixed bitrate, the distortion constraint exerts a greater influence on reconstruction quality than the perceptual constraint: 
as shown in Fig.~\ref{fig:pareto_surface}(b), even a marginal increase in 
$K_D$ from zero produces a substantial uplift in perception. 
Third, as bitrate increases, the maximum FID decreases and its variation range narrows — indicating that realism improves and the sensitivity to perceptual constraints diminishes — while PSNR rises monotonically, confirming that fidelity scales with allocated rate.

Notably, the full RDP trade-off surface is traced solely 
by varying $(K_D, K_P)$ at decoding time — no retraining is required. 
This enables DCIC to generate reconstructions of continuously adjustable fidelity 
and perceptual quality from a single bitstream, 
fundamentally departing from the conventional one-to-one codec paradigm. 
Since $\text{DCIC}_\text{RDP}$, 
$\text{DCIC}_\text{RD}$ and  
$\text{DCIC}_\text{RP}$ 
correspond to the three boundary configurations of the trade-off surface, 
we adopt them as representative operating points for all subsequent comparisons.

\begin{figure}[tb!]
	\centering
	\begin{minipage}[b]{.32\linewidth}
		\centering
		\centerline{\includegraphics[width=2.37in]{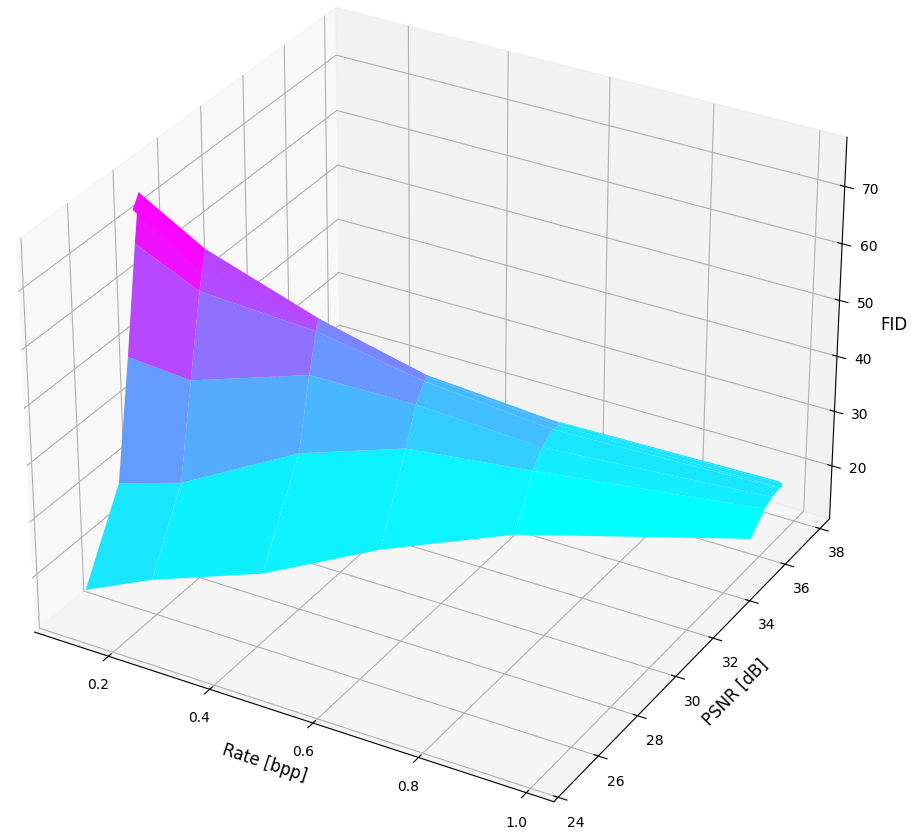}}
		\centerline{(a) RDP Trade-off Surface}\medskip
	\end{minipage}
	\hspace{2.0cm}
	\begin{minipage}[b]{0.32\linewidth}
		\centering
		\centerline{\includegraphics[width=2.34in]{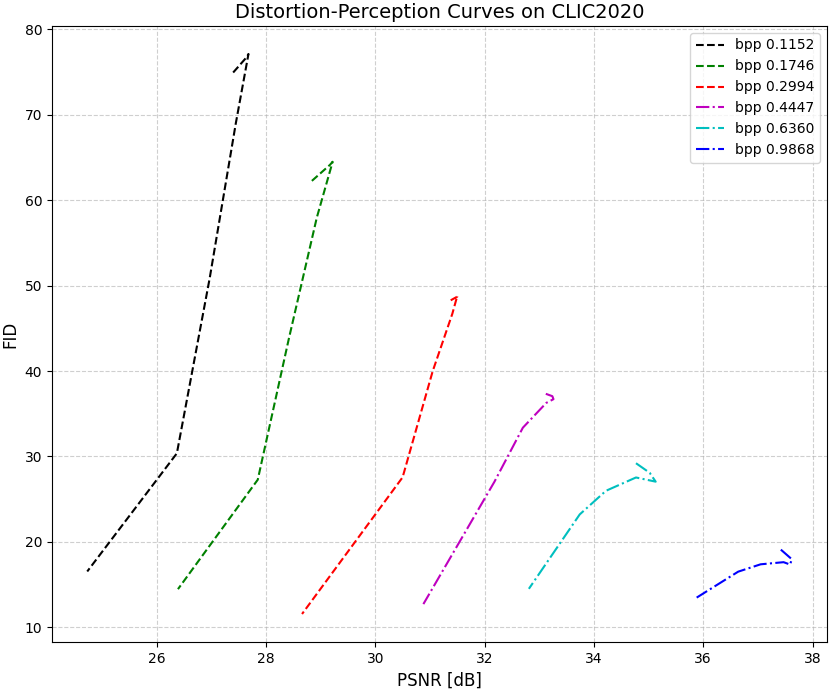}}
		\centerline{(b) Distortion--Perception Pareto Front}\medskip
	\end{minipage}
	\caption{$R(D,P)$ trade-off surface (left) and distortion--perception
		Pareto front (right) of DCIC with Entroformer as the base codec on
		CLIC2020 (0.1152--0.9868~bpp). Seven decoders are obtained by setting
		$(K_D, K_P) \in \{\{1,1\}, \{1,0\}, \{0,1\}, \{1,\frac{1}{2}\},
		\{1,\frac{1}{4}\}, \{1,\frac{1}{8}\}, \{\frac{1}{2},1\}\}$,
		corresponding to $\text{DCIC}_\text{RDP}$, $\text{DCIC}_\text{RD}$,
		$\text{DCIC}_\text{RP}$, $\text{DCIC}_{K_D}(\frac{1}{2})$,
		$\text{DCIC}_{K_D}(\frac{1}{4})$, $\text{DCIC}_{K_D}(\frac{1}{8})$,
		and $\text{DCIC}_{K_P}(\frac{1}{2})$.}	

\label{fig:pareto_surface}
\end{figure}

\subsection{Overall Performance}
\label{sec:results}

We benchmark DCIC against WebP~\cite{google2010webp}, VTM~\cite{bross2021vvc}, 
HiFiC~\cite{Mentzer2020HighFidelityGI}, CDC~\cite{Yang2022LossyIC}, ILLM~\cite{muckley2023illm}, 
IPIC~\cite{xu2024idempotence}, and RDDM~\cite{jiang2025rddm}, 
using BD-PSNR, BD-FID, and BD-LPIPS as evaluation metrics, 
with Hyperprior (Hyper)~\cite{balle2018variational} as the BD anchor. 
Results for the three representative configurations $\ROICRDP$, $\ROICRD$, 
and $\ROICRP$ are reported in Table~\ref{tab:main}.


As shown in Table~\ref{tab:main}, $\text{DCIC}_\text{RDP}$ achieves 
the highest BD-PSNR among all evaluated codecs, 
substantially surpassing both PIC methods (ILLM, IPIC, HiFiC) 
and classical LIC codecs (Hyper, Entroformer), 
while its BD-FID remains comparable to LIC codecs — 
{reflecting a significantly better trade-off between realism and fidelity. 
		Conversely, $\ROICRP$ attains BD-FID 
		comparable to IPIC at the cost of reduced fidelity. 
		Since $\text{DCIC}_\text{RP}$ and $\text{DCIC}_\text{RD}$ share the same architecture 
		and differ only in the perceptual constraint, 
		their performance gap confirms that DCIC subsumes 
		both RD and RP trade-offs within a unified framework.}
	
	Performance gains are most pronounced on CelebA-HQ, where $\text{DCIC}_\text{RDP}$ 
	improves both BD-PSNR and BD-FID simultaneously 
	— an improvement not observed on CLIC2020 or ImageNet-1K. 
	Following Wang et al.~\cite{wang2024semantics}, we attribute this to semantic complexity: 
	CelebA-HQ (faces only) is the least complex of the three datasets, 
	while ImageNet-1K (1,000 categories) is the most, with CLIC2020 in between. 
	Since semantically coherent data better supports optimal RDP, $\text{DCIC}_\text{RDP}$ effectiveness ranks CelebA-HQ > CLIC2020 > ImageNet-1K.

\begin{table}[t]
	\centering
	\caption{BD-metric comparison on CelebA-HQ, CLIC2020, and ImageNet-1K
		(Hyperprior anchor). Lower BD-FID and BD-LPIPS, higher BD-PSNR are
		better. \textbf{Bold} = best per column.}
	\label{tab:main}
	
	\begin{adjustbox}{width=\textwidth} 
		\setlength{\tabcolsep}{1.5pt}
		\small
		\begin{tabular}{lrrr rrr rrr}
			\toprule
			& \multicolumn{3}{c}{CelebA-HQ}
			& \multicolumn{3}{c}{CLIC2020}
			& \multicolumn{3}{c}{ImageNet-1K} \\
			\cmidrule(lr){2-4}\cmidrule(lr){5-7}\cmidrule(lr){8-10}
			Method
			& \scriptsize BD-PSNR$\uparrow$
			& \scriptsize BD-LPIPS$\downarrow$
			& \scriptsize BD-FID$\downarrow$
			& \scriptsize BD-PSNR$\uparrow$
			& \scriptsize BD-LPIPS$\downarrow$
			& \scriptsize BD-FID$\downarrow$
			& \scriptsize BD-PSNR$\uparrow$
			& \scriptsize BD-LPIPS$\downarrow$ 
			& \scriptsize BD-FID$\downarrow$ \\
			\midrule
			Hyperprior~\cite{balle2018variational}
			& 0.00 & 0.00 & 0.00
			& 0.00 & 0.00 & 0.00
			& 0.00 & 0.00 & 0.00 \\
			Entroformer~\cite{qian2022entroformer}
			& 1.3926 & $-$0.031 & $-$13.43
			& 1.2408 & $-$0.044 & $-$5.74
			& 1.0587 & $-$0.042 & $-$7.34 \\
			HiFiC~\cite{mentzer2020hific}
			& $-$2.036 & $-$0.108 & $-$48.35
			& $-$1.621 & \textbf{$-$0.172} & $-$36.16
			& $-$1.418 & \textbf{$-$0.148} & $-$44.52 \\
			CDC~\cite{yang2023cdc}
			& $-$8.014 & $-$0.060 & $-$43.80
			& $-$7.043 & $-$0.084 & $-$38.31
			& $-$6.416 & $-$0.084 & $-$41.75 \\
			ILLM~\cite{muckley2023illm}
			& $-$1.234 & \textbf{$-$0.109} & $-$50.58
			& $-$0.480 & $-$0.155 & \textbf{$-$48.22}
			& $-$0.596 & $-$0.181 & $-$42.95 \\
			IPIC(Hyp.)~\cite{xu2024idempotence}
			& $-$2.225 & $-$0.086 & $-$54.14
			& $-$2.920 & $-$0.056 & $-$44.52
			& $-$2.648 & $-$0.058 & $-$52.12 \\
			IPIC(ELIC)~\cite{xu2024idempotence}
			& $-$0.986 & $-$0.099 & \textbf{$-$54.89}
			& $-$1.635 & $-$0.079 & $-$46.52
			& $-$1.492 & $-$0.106 & \textbf{$-$55.18}\\
			WebP~\cite{google2010webp}
			& $-$2.525 & $-$0.004 & 15.00
			& $-$2.374 & 0.017 & 28.80
			& $-$1.709 & $-$0.006 & 8.27 \\
			VTM~\cite{bross2021vvc}
			& 0.7495 & $-$0.031 & $-$14.22
			& 1.0370 & $-$0.048 & $-$12.21
			& 0.9018 & $-$0.048 & $-$13.11 \\
			\midrule
			$\ROICRP$ (ours)
			& $-$1.3577 & $-$0.095 & $-$54.03
			& $-$1.4108 & $-$0.076 & $-$46.19
			& $-$2.0331 & $-$0.067 & $-$52.72 \\
			$\ROICRD$ (ours)
			& 1.3347 & $-$0.037 & $-$25.31
			& 1.1617 & $-$0.019 & $-$5.58
			& 1.0623 & $-$0.031 & $-$5.41\\
			$\ROICRDP$ (ours)
			& \textbf{1.4658} & $-$0.040 & $-$26.45
			& \textbf{1.3671} & $-$0.041 & $-$5.13
			& \textbf{1.1481} & $-$0.034 & $-$5.02 \\
			\bottomrule
		\end{tabular}
	\end{adjustbox}
\end{table}

\subsection{Ablation: Constraint Contributions}
\label{sec:ablation}

Table~\ref{tab:ablation} isolates the effect of $\cD$ and $\cP$.
Without any constraint, the unconditioned diffusion model yields
PSNR~$=9.90$~dB and MSE-Ratio~$\approx 0$.
Adding $\cD$ alone ($\ROICRD$) recovers PSNR to 38.12~dB (Ratio~$=1.01$),
matching Entroformer.
Adding $\cP$ alone ($\ROICRP$) yields PSNR~$=33.99$~dB but
Ratio~$=0.39$, demonstrating strong perceptual guidance at moderate
fidelity cost.
$\ROICRDP$ achieves the best of both: PSNR~$=38.34$~dB and
Ratio~$=1.06$, surpassing the base codec on all fidelity metrics. 
Additional ablation studies are provided in Supplementary E.

\begin{table}[thb]
	\centering
	\caption{
		Ablation of distortion $\cD$ and perceptual $\cP$ constraints in DCIC
		(Entroformer, $\lambda\!=\!0.02$).
	}
	\label{tab:ablation}
	\setlength{\tabcolsep}{6pt}
	\small
	\begin{tabular}{lcc cccc}
		\toprule
		Method & $\cD$ & $\cP$ & PSNR$\uparrow$ & MS-SSIM$\uparrow$ & MSE$\downarrow$ & Ratio$\downarrow$ \\
		\midrule
		Entroformer~\cite{qian2022entroformer} & $\times$ & $\times$ & 38.09 & 0.9889 & 0.000621 & 1.00 \\
		DM (uncond.)                          & $\times$ & $\times$ & 9.903 & 0.1777 & 0.409065 & 0.00 \\
		$\ROICRD$                             & \checkmark & $\times$ & 38.12 & 0.9890 & 0.000616 & 1.01 \\
		$\ROICRP$                             & $\times$ & \checkmark & 33.99 & 0.9749 & 0.001594 & 0.39 \\
		$\ROICRDP$                            & \checkmark & \checkmark
		& \textbf{38.34} & \textbf{0.9892} & \textbf{0.000587} & \textbf{1.06} \\
		\bottomrule
	\end{tabular}
\end{table}

\subsection{Computational Complexity}
\label{sec:complexity}

HiFiC and ILLM require $\sim$1 week of training per rate point.
DCIC, IPIC, and DDIM require no training, offering greater
flexibility.
Inference takes $\sim$63\,s for DCIC vs.\ $\sim$60\,s for IPIC, with
overhead attributable to per-step base codec gradient computation.
Unlike one-to-one codecs, DCIC generates $N$ reconstructions from a
single bitstream by varying $(K_D,K_P)$, subsuming multiple
traditional codec behaviors within one framework.
In addition, since $\ROICRDP$ enforces both constraints simultaneously, 
it incurs a marginally higher computational cost than $\ROICRD$ and $\ROICRP$, 
each of which applies only a single constraint.

\begin{table}[!t]
	\centering
	\begin{threeparttable}
		\caption{
			Computational complexity of DCIC and other traditional RP methods. Here, $K$ denotes the number of supported bitrates, with $K=3$ for HiFiC and $K=6$ for ILLM.
		}
		\label{tab:complexity}
		\small
		\begin{tabular}{lccccc}
			\toprule
			\multirow{2}{*}{Methods} & \multirow{2}{*}{Number of models}  & {Number of}   & {Number of} & \multirow{2}{*}{Train} & \multirow{2}{*}{Inference}\\ 
			&  & {reconstructions} & {time-steps}  &  &\\
			\midrule
			{HiFiC, ILLM} & {\(K\)} & \multirow{2}{*}{1}  & {-} & \( \sim K \)(Weeks)  & {\( \sim 0.1s \)} \\
			DDIM & 1 & 1 & 250 & 0 & \( \sim 50s \) \\
			IPIC & 1  & 1 & 1000 & 0 & \( \sim 60s \) \\
			DCIC~(ours) & 1  & $\boldsymbol{N}$ & 250 & 0 & \( \sim 63s \) \\
			\bottomrule
		\end{tabular}
	\end{threeparttable}
\end{table}

\section{Discussion}
\noindent\textbf{Relationship to IPIC and RDDM.}
The three methods pursue {entirely different optimisation
	objectives}: IPIC and RDDM each solve a scalar problem targeting a
single output (RP and RD trade-off, respectively), whereas DCIC solves
a \emph{bi-objective} problem targeting the full Pareto frontier of the
$(D,P)$ plane, producing \emph{multiple} outputs of continuously
adjustable fidelity--realism from a single bitstream.
As shown in Table~\ref{tab:ablation}, $\text{DCIC}_\text{RP}$
($K_D\!=\!0$) and $\text{DCIC}_\text{RDP}$ ($K_P\!=\!K_P\!=\!1$) subsume IPIC
and RDDM as the two boundary curves of this frontier, confirming that
the prior methods are special cases of DCIC rather than alternatives.
Moreover, as $K_P \to 0$, $\text{DCIC}_\text{RD}$ progressively
approximates the base codec, establishing a smooth continuum from
pure codec behaviour to full RDP-optimal decoding along the distortion
axis.
The dual-constraint objective $J^{(t)}_{\mathrm{DP}}$ is derived from
a single unified log-posterior under the Markov chain
$x\!\to\!\hat{x}\!\to\!\tilde{x}$, and is the only framework formally
connected to the operational $R_C(D,P)$ function. 
Common randomness is realised by applying the \emph{same} $\gc$ to
$\tilde{x}_t$ identically at encoder and decoder, requiring no
additional transmitted bits.


\begin{table}[!t]
	\centering
	\begin{threeparttable}
		\caption{Relationship between DCIC and IPIC, RDDM, and the base codec.
			}
		\label{tab:novelty}
			\begin{tabular}{lcccc}
			\toprule
			\textbf{Dimension} & IPIC & RDDM & Base Codec & {DCIC} \\
			\midrule
			Optimization target & \makecell{RP-Tradeoffs \\ (perception)} &  \makecell{RD-Tradeoffs \\ (distortion)}  & \makecell{RD-Tradeoffs \\ (distortion)} & \makecell{RDP-Tradeoffs \\ (Pareto front)} \\
			Outputs~/~Bitstream & One & One & One & {Multiple} \\
			Constraint type & \makecell{Iempotence } & \makecell{$\cD + \cP$$^\dagger$ } & -- & \makecell{$\cD\cup\cP$ } \\ 
			Common randomness & \texttimes & \texttimes & \texttimes & \checkmark \\
			$R_C(D,P)$ connection & \texttimes & \texttimes & \texttimes & \checkmark \\
			Subsumed by DCIC when & $K_D\!=\!0$ & $K_D\!=\!K_P\!=\!1$ &  $K_P\!=\!0$ & -- \\
			\bottomrule
		\end{tabular}
		\footnotesize
		\begin{tablenotes}
		\item[$\dagger$] 
		Within RDDM, during its first 200 iterations, $\lambda_1$ converges to its optimal value 
		while $\lambda_2 \approx 0$; in the final 50 iterations, this pattern reverses, 
		with $\lambda_2$ converging while $\lambda_1 \approx 0$. 
		 $\lambda_1$ and $\lambda_2$ are corresponding to $\cD$ and $\cP$, respectively. 
		\end{tablenotes}
	\end{threeparttable}
\end{table}
	

\noindent\textbf{Limitations and future work.}
The $\sim$63,s inference cost of 250-step diffusion sampling is a practical bottleneck to DCIC. Accelerated samplers — such as consistency models, DEIS, or the recently proposed Drifting Models~\cite{deng2026drifting}, which achieve high-quality generation in a single step — could substantially reduce this overhead.
%
The approximation $\partial\gc/\partial\tilde{x}_0\approx 1$ degrades
at very low bitrates, motivating bitrate-aware gradient corrections.
Semantic-aware weight functions $\lambda_D(t),\lambda_P(t)$ conditioned
on CLIP features could close the performance gap on semantically complex
datasets.
Extension to video compression within the RDP framework is a promising
future direction.

\section{Conclusion}
\label{sec:conclusion}
We presented DCIC, a dual-constrained diffusion decoding framework that
operationalises the full distortion--perception Pareto frontier of
neural image compression at a fixed rate.
By jointly imposing a distortion constraint $\mathcal{C}_D$ and an
idempotence constraint $\mathcal{C}_P$ on the diffusion reverse process,
DCIC derives a unified bi-objective iterative algorithm grounded in the
operational $R_C(D,P)$ function. 
Consistent noise injection via the base codec $\gc$ realises common
randomness across encoding and decoding without additional rate
overhead, while idempotence ensures reconstructions remain conditioned
on the encoder output.
Attenuation factors $(K_D, K_P)\in[0,1]^2$ enable continuous
navigation of the $(D,P)$ Pareto frontier from a single bitstream,
subsuming $\text{DCIC}_\text{RD}$ ($K_P\!=\!0$) and
$\text{DCIC}_\text{RP}$ ($K_D\!=\!0$) as boundary curves and
$\text{DCIC}_\text{RDP}$ ($K_D\!=\!K_P\!=\!1$) as the optimal
interior operating point.
Comprehensive evaluations across CNN, Transformer, and hybrid LIC
architectures on CelebA-HQ, CLIC2020, and ImageNet-1K confirm
state-of-the-art BD-PSNR and competitive BD-FID, with strong
generalizability across base codec architectures.

\noindent\textbf{Broader Impacts.} DCIC improves bandwidth efficiency
and perceptual quality for image transmission in resource-constrained
settings. The ability to generate perceptually realistic
reconstructions may carry misinformation risks; see Supplementary~H
for a full discussion.


\bibliographystyle{unsrtnat}

\newpage
\appendix
\renewcommand{\thesection}{\Alph{section}}
\renewcommand{\thesubsection}{\Alph{section}.\arabic{subsection}}
\renewcommand{\thetable}{S\arabic{table}}
\renewcommand{\thefigure}{S\arabic{figure}}
\renewcommand{\theequation}{S\arabic{equation}}

\section{Technical appendices and supplementary material}


This document provides additional details that support the main paper but
could not be included within the page limit.
It is organized as follows:
Sec.~\ref{sec:derivations} gives the full mathematical derivations.
Sec.~\ref{sec:extended_exp} provides extended experimental results.
Sec.~\ref{sec:algorithm} gives the complete algorithm pseudocode.
Sec.~\ref{sec:ablations} covers additional ablation studies.
Sec.~\ref{sec:impl} details the implementation and reproducibility information.
Sec.~\ref{sec:limitations} extends the limitations discussion. 
Sec.~\ref{sec:broader_impacts} details the broader impacts.

\section{Full Mathematical Derivations}
\label{sec:derivations}

\subsection{Per-Step RDP Objective (Eq.~7, Main Paper)}
\label{sec:deriv_obj}

Starting from the negative log-likelihood of the reverse step under
constraints $\cD$ and $\cP$:
\begin{align}
	J^{(t)}_{\mathrm{DP}}
	&= -\log p_\theta(x_{t-1}\mid x_t,\cD,\cP) \notag\\
	&= -\log p_\theta(\cD,\cP\mid x_{t-1},x_t)
	-\log p_\theta(x_{t-1}\mid x_t) + K'. \label{eq:s1}
\end{align}
By Bayes' theorem and the Markov structure of the reverse diffusion
process, we factorize the joint constraint term:
\begin{equation}
	= -\log p_\theta(\cP\mid x_{t-1})
	-\log p_\theta(\cD\mid x_{t-1},\cP)
	-\log p_\theta(x_{t-1}\mid x_t) + K'.
	\label{eq:s2}
\end{equation}
Applying the Markov property of the reverse sampling process (which
renders $\cD$ and $\cP$ conditionally independent given $x_{t-1}$):
\begin{equation}
	\approx -\log p_\theta(\cP\mid x_{t-1})
	-\log p_\theta(\cD\mid x_{t-1})
	-\log p_\theta(x_{t-1}\mid x_t) + K.
	\label{eq:s3}
\end{equation}
Substituting Gaussian likelihood models for each constraint term:
\begin{itemize}
	\item $\cP$ (idempotence): $-\log p_\theta(\cP\mid x_{t-1})
	\propto \tfrac{1}{2\xi_t^2}\|\bxhat-\gc(\tilde{x}_0)\|^2$
	\item $\cD$ (distortion): $-\log p_\theta(\cD\mid x_{t-1})
	\propto \tfrac{1}{2\xi_t^2}\|\bxhat-\tilde{x}_0\|^2$
	\item Denoising prior (standard DDPM reverse step):
	$-\log p_\theta(x_{t-1}\mid x_t)
	\propto \tfrac{1}{2\sigma_t^2}\|\tilde{x}_t-\tilde\mu_t\|^2$
\end{itemize}
Combining yields Eq.~8 of the main paper:
\begin{equation}
	J^{(t)}_{\mathrm{DP}} \approx
	\frac{1}{2\xi_t^2}\!\left[
	\|\bxhat-\gc(\tilde{x}_0)\|^2 + \|\bxhat-\tilde{x}_0\|^2
	\right]
	+\frac{1}{2\sigma_t^2}\|\tilde{x}_t-\tilde\mu_t\|^2+K.
	\label{eq:jt_supp}
\end{equation}
The two bracketed terms correspond respectively to the idempotence
residual (perception constraint gradient) and the fidelity residual
(distortion constraint gradient).

\subsection{Gradient Derivation (Eq.~8, Main Paper)}
\label{sec:deriv_grad}

Differentiating Eq.~\eqref{eq:jt_supp} w.r.t.\ $\tilde{x}_t$:
\begin{align}
	\nabla_{\tilde{x}_t} J^{(t)}_{\mathrm{DP}}
	&= \frac{1}{\xi_t^2}[\nabla_{\tilde{x}_t}\gc(\tilde{x}_0)]
	[\bxhat-\gc(\tilde{x}_0)]
	+ \frac{1}{\xi_t^2}[\bxhat-\tilde{x}_0]
	+ \frac{1}{\sigma_t^2}(\tilde{x}_t-\mu_t).
	\label{eq:grad_s1}
\end{align}
Applying the chain rule through the DDIM one-step approximation
$f_\theta(\tilde{x}_t)$ (Eq.~8 of the main paper):
\begin{align}
	&= \frac{1}{\xi_t^2}\cdot
	\frac{\partial\gc(\tilde{x}_0)}{\partial\tilde{x}_0}\cdot
	\frac{\partial f_\theta(\tilde{x}_t)}{\partial\tilde{x}_t}
	\cdot[\bxhat-\gc(\tilde{x}_0)] \notag\\
	&\quad+ \frac{1}{\xi_t^2}\cdot
	\frac{\partial f_\theta(\tilde{x}_t)}{\partial\tilde{x}_t}
	\cdot[\bxhat-\tilde{x}_0]
	+ \frac{1}{\sigma_t^2}(\tilde{x}_t-\mu_t),
	\label{eq:grad_s2}
\end{align}
where
\begin{equation}
	\frac{\partial f_\theta(\tilde{x}_t)}{\partial\tilde{x}_t}
	= \frac{1-\sqrt{1-\bar\alpha_t}\,\nabla_{\tilde{x}_t}\epsth(\tilde{x}_t,t)}
	{\sqrt{\bar\alpha_t}}.
	\label{eq:ddim_jac}
\end{equation}

\noindent\textbf{Approximation $\partial\gc/\partial\tilde{x}_0\approx 1$.}
At sufficiently high bitrates, the codec output closely tracks the input,
so variations in $\tilde{x}_0$ produce proportional changes in $\gc(\tilde{x}_0)$
with unit gain.
Substituting and factoring:
\begin{equation}
	\nabla_{\tilde{x}_t} J^{(t)}_{\mathrm{DP}}
	\approx
	\frac{2}{\xi_t^2}\cdot
	\frac{\partial f_\theta(\tilde{x}_t)}{\partial\tilde{x}_t}
	\cdot\!\left[\bxhat-\frac{\gc(\tilde{x}_0)+\tilde{x}_0}{2}\right]
	+\frac{1}{\sigma_t^2}(\tilde{x}_t-\mu_t).
	\label{eq:grad_final}
\end{equation}
This approximation is the primary source of DCIC's performance
degradation at low bitrates (where $\partial\gc/\partial\tilde{x}_0\ll 1$),
and is validated empirically in Sec.~\ref{sec:per_bitrate}.

\subsection{Optimality Conditions and Convergence Analysis}
\label{sec:convergence}

Because $\JRDP$ is convex in $\tilde{x}_t$, gradient descent globally
converges.
However, the implicit approximation $D\approx 0$ in Eq.~\eqref{eq:jt_supp}
means that the optimal RDP point does \emph{not} coincide with the
minimum of $\JRDP$.
Specifically, substituting the source image $x$ into
Eq.~\eqref{eq:grad_final} shows that the distortion term $(x-\hat{x})$ 
does not vanish at $\tilde{x}_0=x$.
Consequently:
\begin{itemize}
	\item Gradient descent initially improves reconstruction quality as
	$\tilde{x}_0\to x$, but eventually overshoots and degrades quality.
	\item The gamma learning rate $\eta(t)$ and weighting functions
	$\lambda_D(t),\lambda_P(t)$ are jointly designed to ensure effective
	termination near the quality peak, rather than at the gradient minimum.
	\item The perception weight $\lambda_P(t)$ decays more rapidly than
	$\lambda_D(t)$ because the idempotence gradient must approach zero
	(to satisfy $\cP$ exactly), while the distortion gradient only
	needs to remain within $2\Delta^*$ (to satisfy $\cD$).
\end{itemize}

\section{Extended Experimental Results}
\label{sec:extended_exp}

\subsection{Full RDP Surface Analysis}
\label{sec:rdp_surface_supp}

\begin{figure}[tbh!]
	\centering
	\begin{minipage}[b]{0.35\linewidth}
		\centering
		\centerline{\includegraphics[width=2.45in]{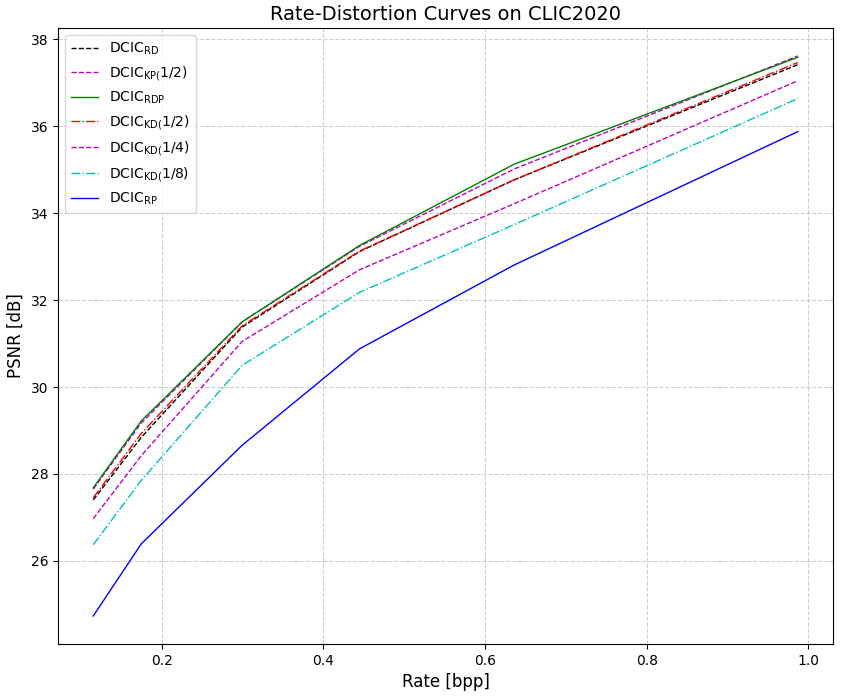}}
		\centerline{(b) Rate-Distortion Curves}\medskip
	\end{minipage}
	\hspace{1.5cm}
	\begin{minipage}[b]{0.35\linewidth}
		\centering
		\centerline{\includegraphics[width=2.45in]{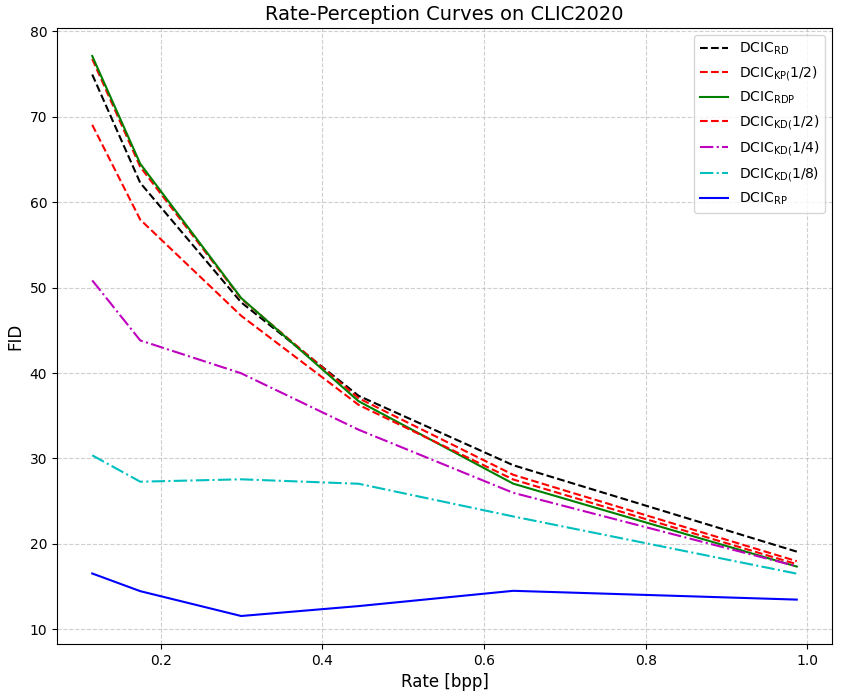}}
		\centerline{(a) Rate-Perception Curves}\medskip
	\end{minipage}
	
	\caption{
		Rate--perception (left) and distortion--perception (right) curves of DCIC with Entroformer as the base codec on CLIC2020 (0.1152--0.9868~bpp), complementing the rate--distortion curves in Fig.~3 of the main paper to fully characterize the RDP trade-off surface.
	} 
	\label{fig:rdp_surface}
\end{figure}

The $R(D,P)$ surface is constructed from seven DCIC configurations on
CLIC2020 using Entroformer as base codec:
$\ROICRDP$, $\ROICRD$, $\ROICRP$, and
$\text{DCIC}_{K_D}(1/2)$, $\text{DCIC}_{K_D}(1/4)$,
$\text{DCIC}_{K_D}(1/8)$, and $\text{DCIC}_{K_P}(1/2)$,
across bitrates $0.1152$--$0.9868$~bpp.
Three key properties emerge consistent with RDP theory:

\begin{enumerate}
	\item \textbf{Monotone rate effect.}
	As bitrate increases, both minimum PSNR and maximum FID improve monotonically, 
	while the achievable FID range narrows — indicating that the distortion--perception trade-off is more critical at low bitrates.
	
	\item \textbf{Dominance of distortion constraint.}
	As shown in Fig.~S1(b), at any fixed bitrate, small variations in distortion produce substantial changes in perception, 
	whereas variations in perception exert little influence on distortion 
	— an asymmetry that is most pronounced at low bitrates. 
	This justifies the finer discretization of the distortion attenuation factor $K_D$
	relative to the perceptual attenuation factor $K_P$ in DCIC.
	
	\item \textbf{High-bitrate saturation.}
	Above $\sim$0.7~bpp, the surface flattens in the perception direction, 
	indicating that distortion constraint variations exert negligible influence on perceptual quality 
	— suggesting that the codec naturally approaches perceptual optimality without explicit guidance.
\end{enumerate}

Fig.~\ref{fig:rdp_surface} presents the rate--perception and distortion--perception curves for all seven DCIC decoders, 
complementing the RDP surface in Fig.~3 of the main paper. 
As shown in Fig.~\ref{fig:rdp_surface}(a), the perceptual quality of all seven decoders converges at high bitrates, 
indicating that the perceptual constraint exerts diminishing influence as rate increases. 
This is further corroborated by Fig.~\ref{fig:rdp_surface}(b), 
where the distortion--perception trade-off range narrows monotonically with bitrate. 
Additionally, the distortion--perception curve in Fig.~\ref{fig:rdp_surface}(b) exhibits a clear inflection point, 
consistent with the theoretical convexity of $R(D,P)$.


\begin{figure}[!tb]
	\centering
	\begin{minipage}[b]{.35\linewidth}
		\centering
		\centerline{\includegraphics[width=2.44in]{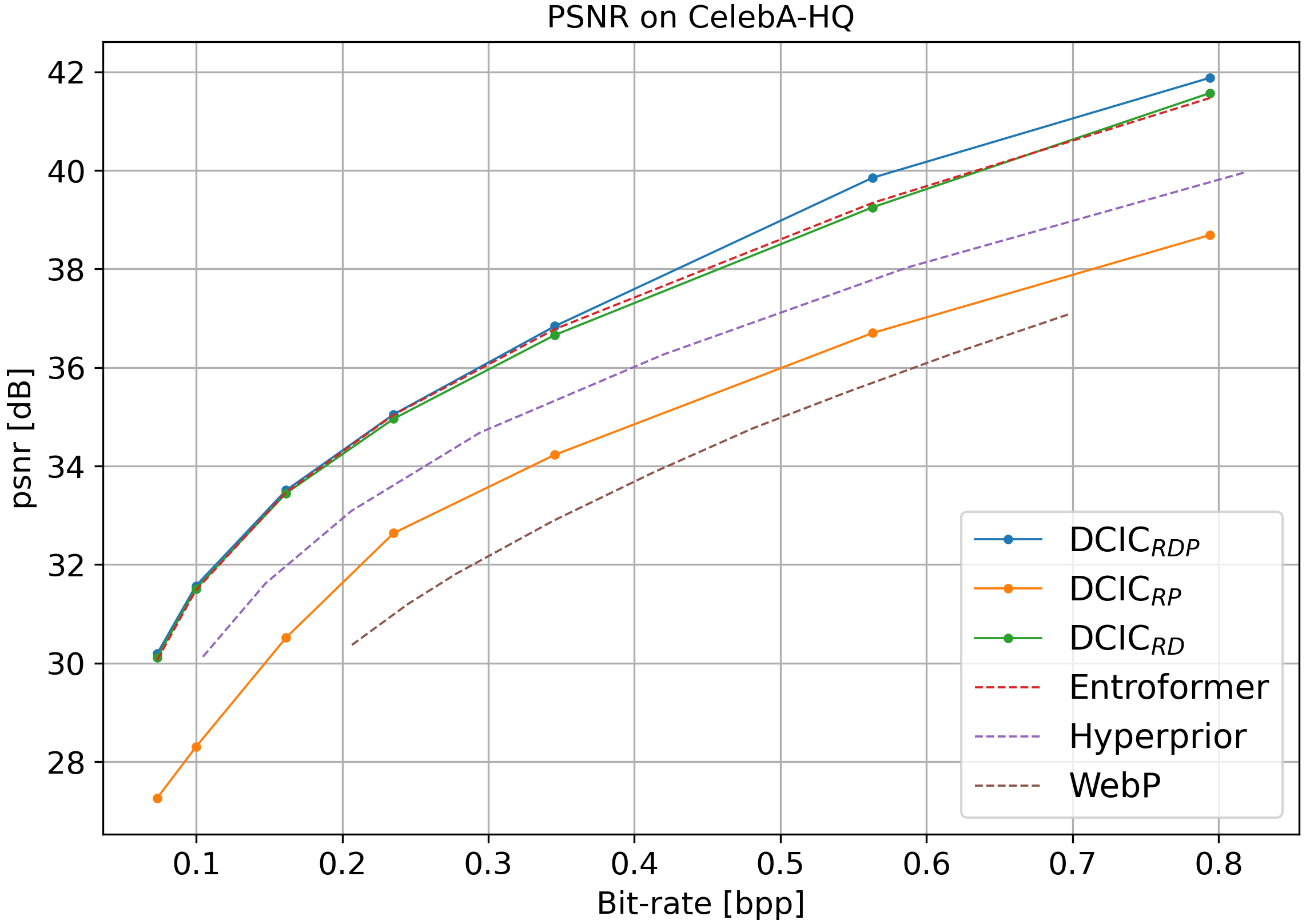}}
		\centerline{(a) PSNR on CelebA-HQ}\medskip
	\end{minipage}
	\hspace{2.0cm}
	\begin{minipage}[b]{.35\linewidth}
		\centering
		\centerline{\includegraphics[width=2.44in]{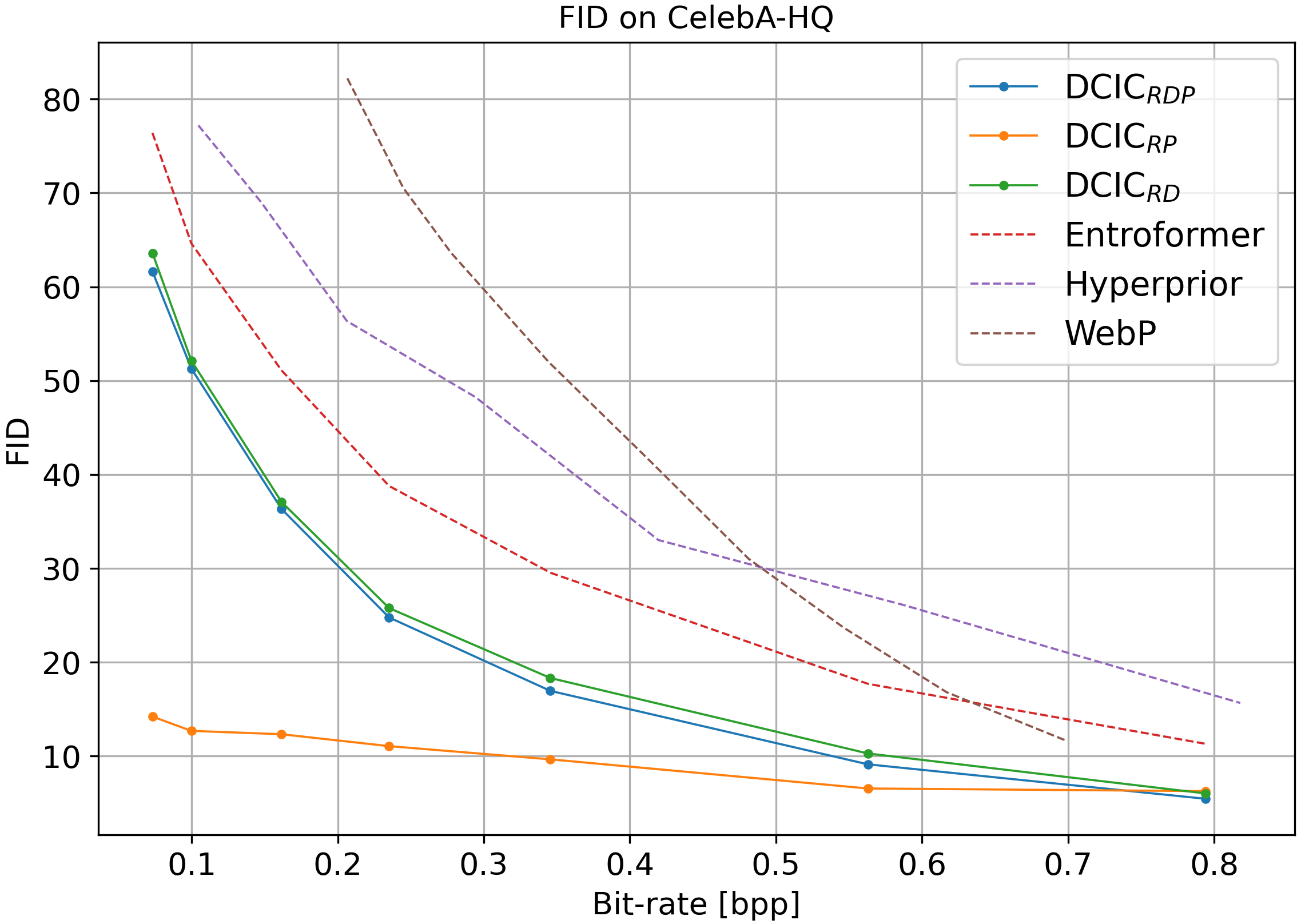}}
		\centerline{(b) FID on CelebA-HQ}\medskip
	\end{minipage}
	\hfill
	\begin{minipage}[b]{0.35\linewidth}
		\centering
		\centerline{\includegraphics[width=2.44in]{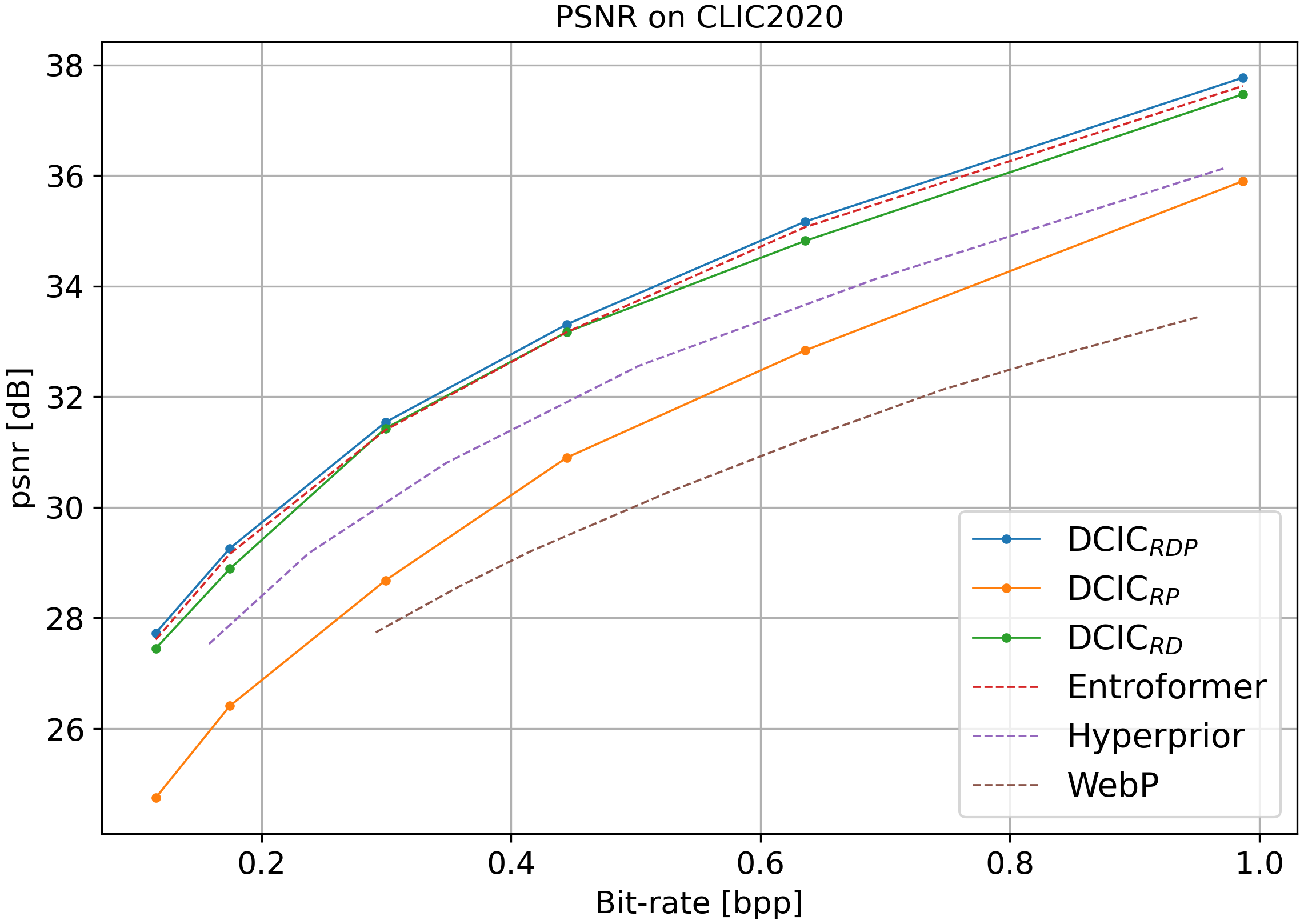}}
		\centerline{(c) PSNR on CLIC2020}\medskip
	\end{minipage}
	\hspace{2.0cm}
	\begin{minipage}[b]{0.35\linewidth}
	\centering
	\centerline{\includegraphics[width=2.44in]{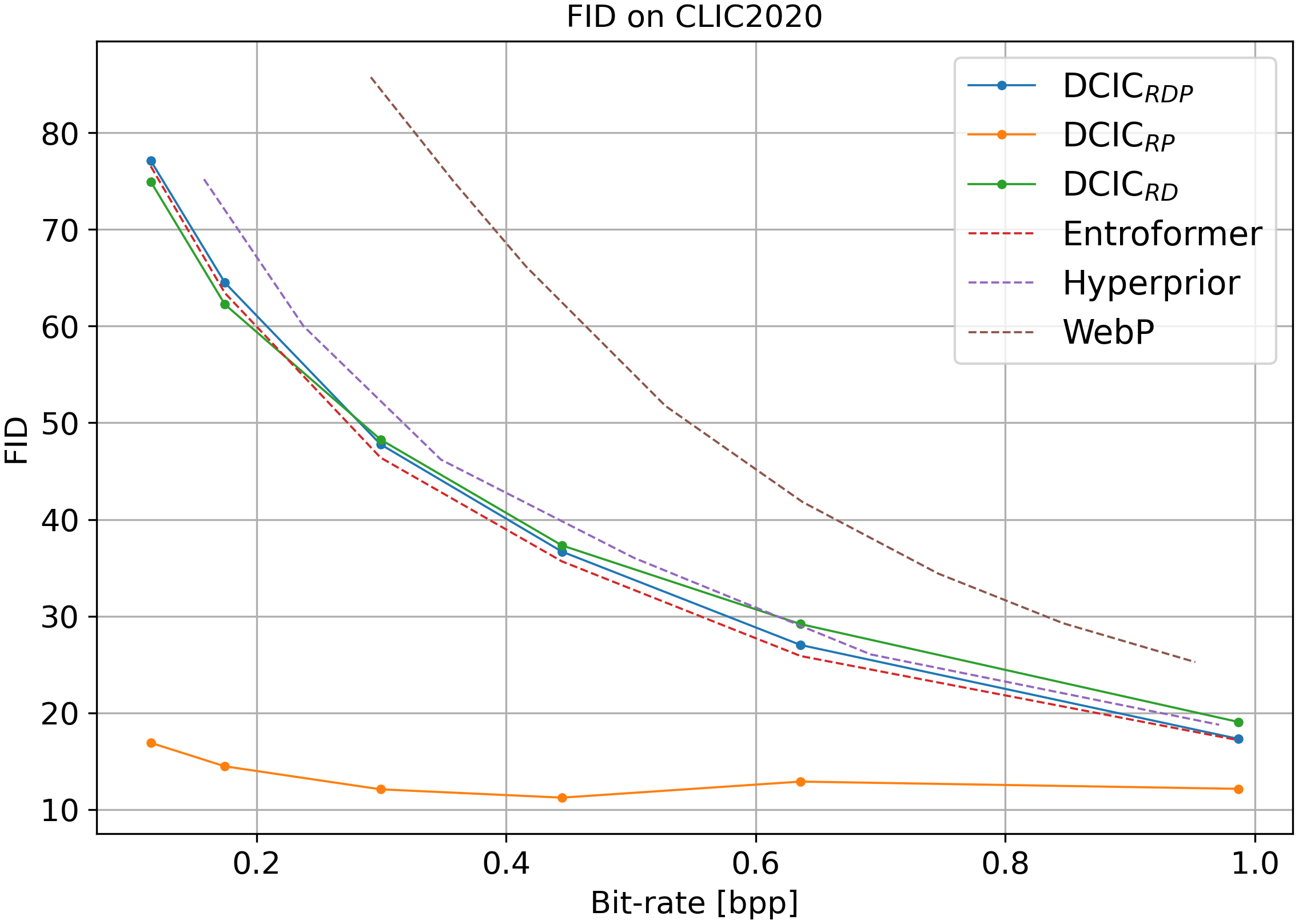}}
	\centerline{(d) FID on CLIC2020}\medskip
	\end{minipage}
	
		\hfill

	\begin{minipage}[b]{0.35\linewidth}
		\centering
		\centerline{\includegraphics[width=2.44in]{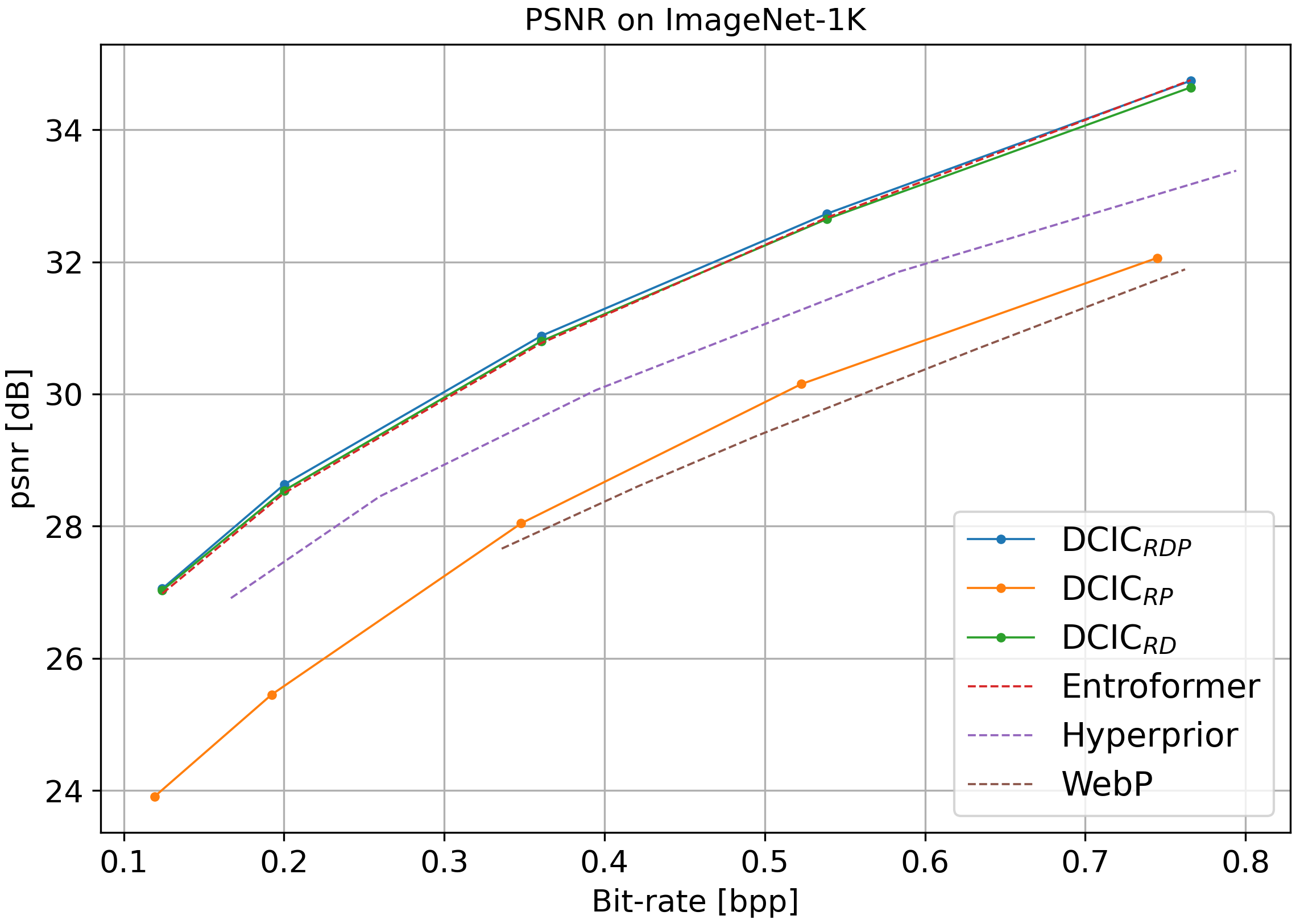}}
		\centerline{(e) PSNR on Imagenet-1K}\medskip
	\end{minipage}
	\hspace{2.0cm}
		\begin{minipage}[b]{0.35\linewidth}
		\centering
		\centerline{\includegraphics[width=2.44in]{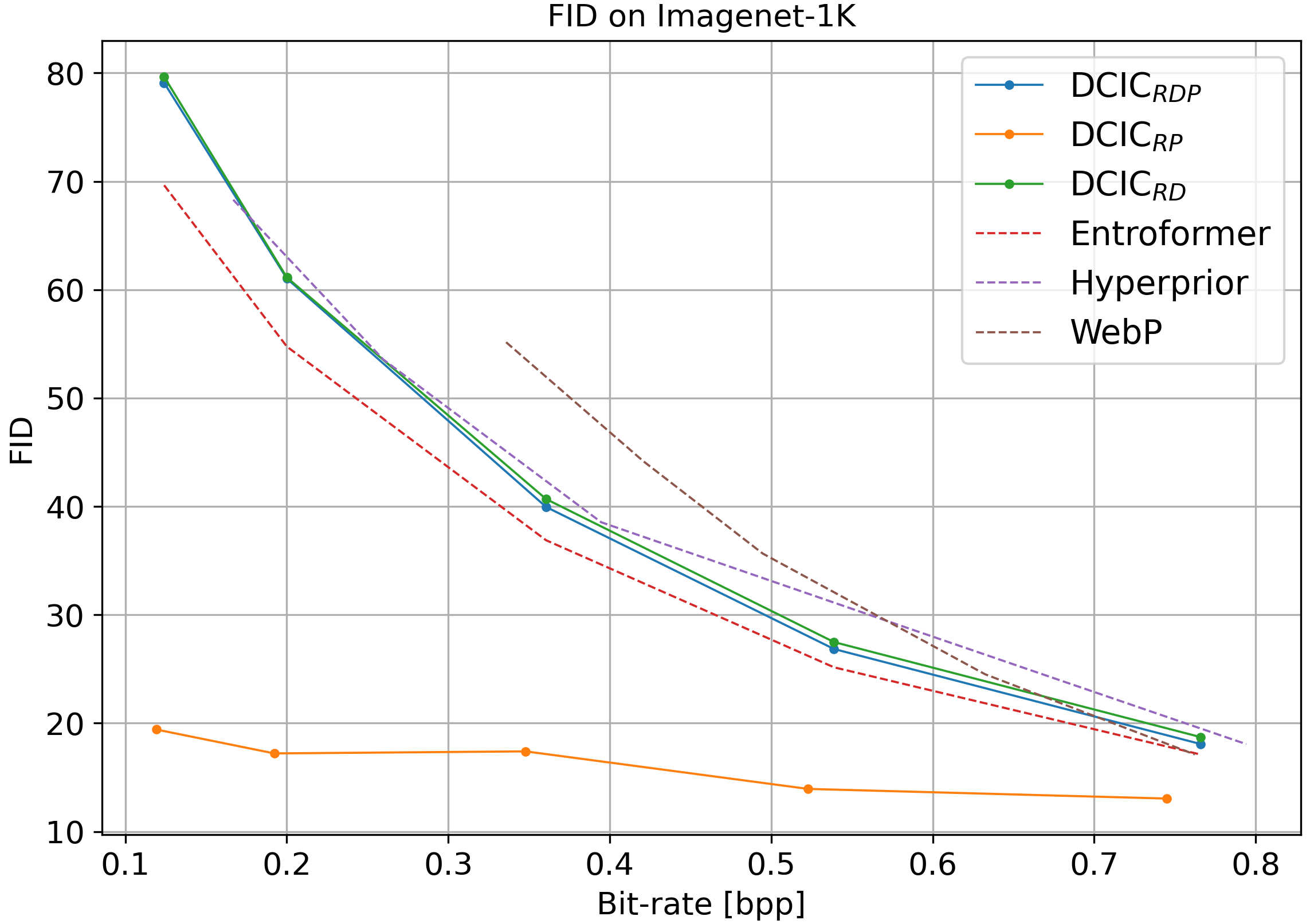}}
		\centerline{(f) FID on ImageNet-1K}\medskip
	\end{minipage}
	\caption{
		Rate--distortion and rate--perception curves of 
		$\text{DCIC}_\text{RDP}$, $\text{DCIC}_\text{RD}$, $\text{DCIC}_\text{RP}$, 
		Entroformer, Hyperprior, and WebP, measured by PSNR and FID respectively. 
		(a), (c), (e): rate--distortion on CelebA-HQ, CLIC2020, and ImageNet-1K; 
		(d), (e), (f): rate--perception on CelebA-HQ, CLIC2020, and ImageNet-1K.
	} 
\label{fig:detailed_cureves}
\end{figure}

\begin{figure}[!tb]
\centering

\includegraphics[width=\linewidth]{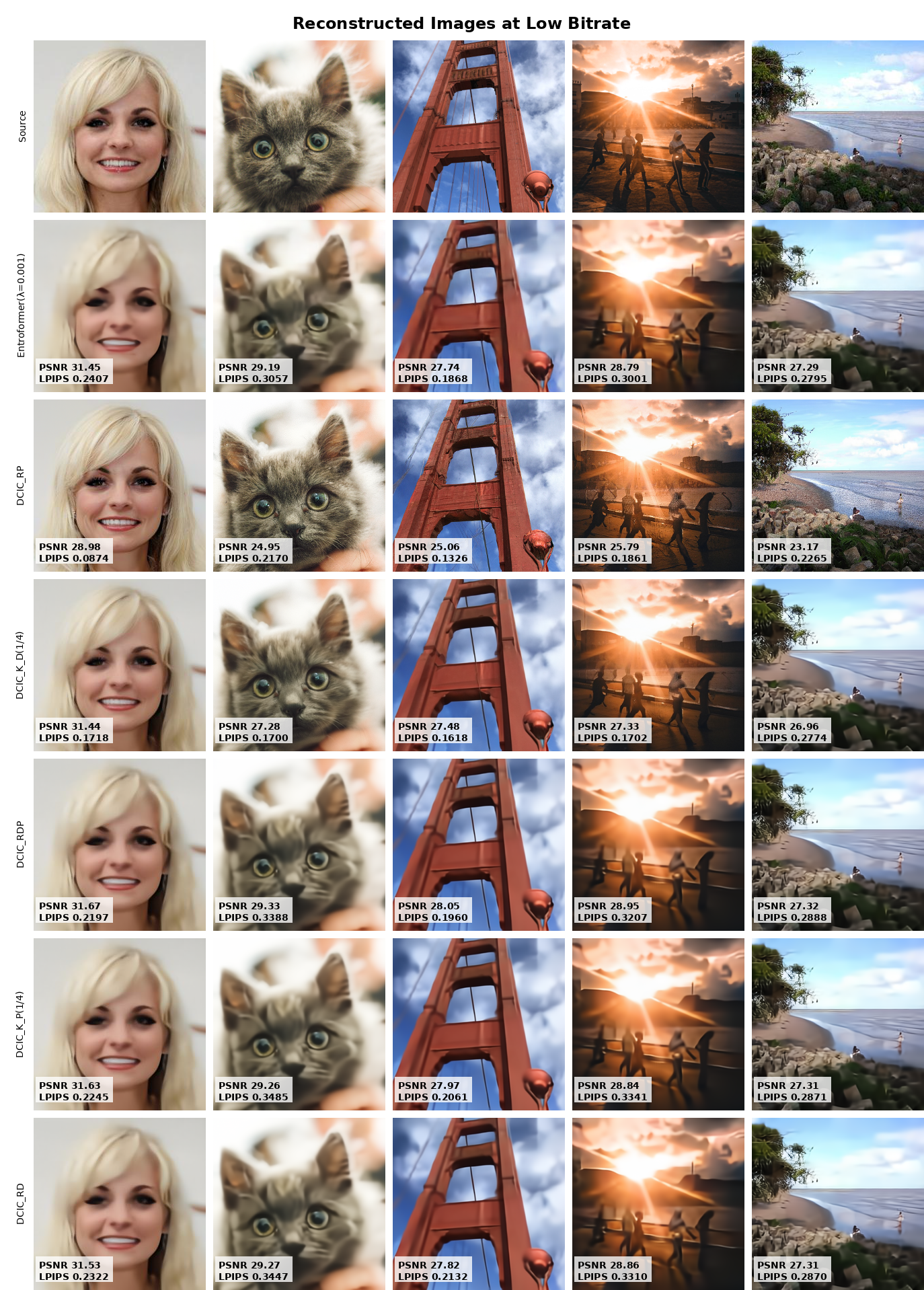}

\caption{
	Visual comparison of reconstructed and restored images at low bitrate ($\lambda=0.001$).  
	The first row shows the source images, the second row presents reconstructions from Entroformer, 
	and rows 3–7 display restorations generated by DCIC variants under different attenuation factors. 
	Among the DCIC variants, DCIC$_{\text{RDP}}$ achieves the best fidelity (highest PSNR), 
	DCIC$_{\text{RD}}$ produces results most similar to the Entroformer output 
	(with closely aligned PSNR and LPIPS), and DCIC$_{\text{RP}}$ 
	attains the highest perceptual quality (lowest LPIPS).
}
\label{fig:demons}
\end{figure}

\begin{figure}[!tb]
\centering

\includegraphics[width=\linewidth]{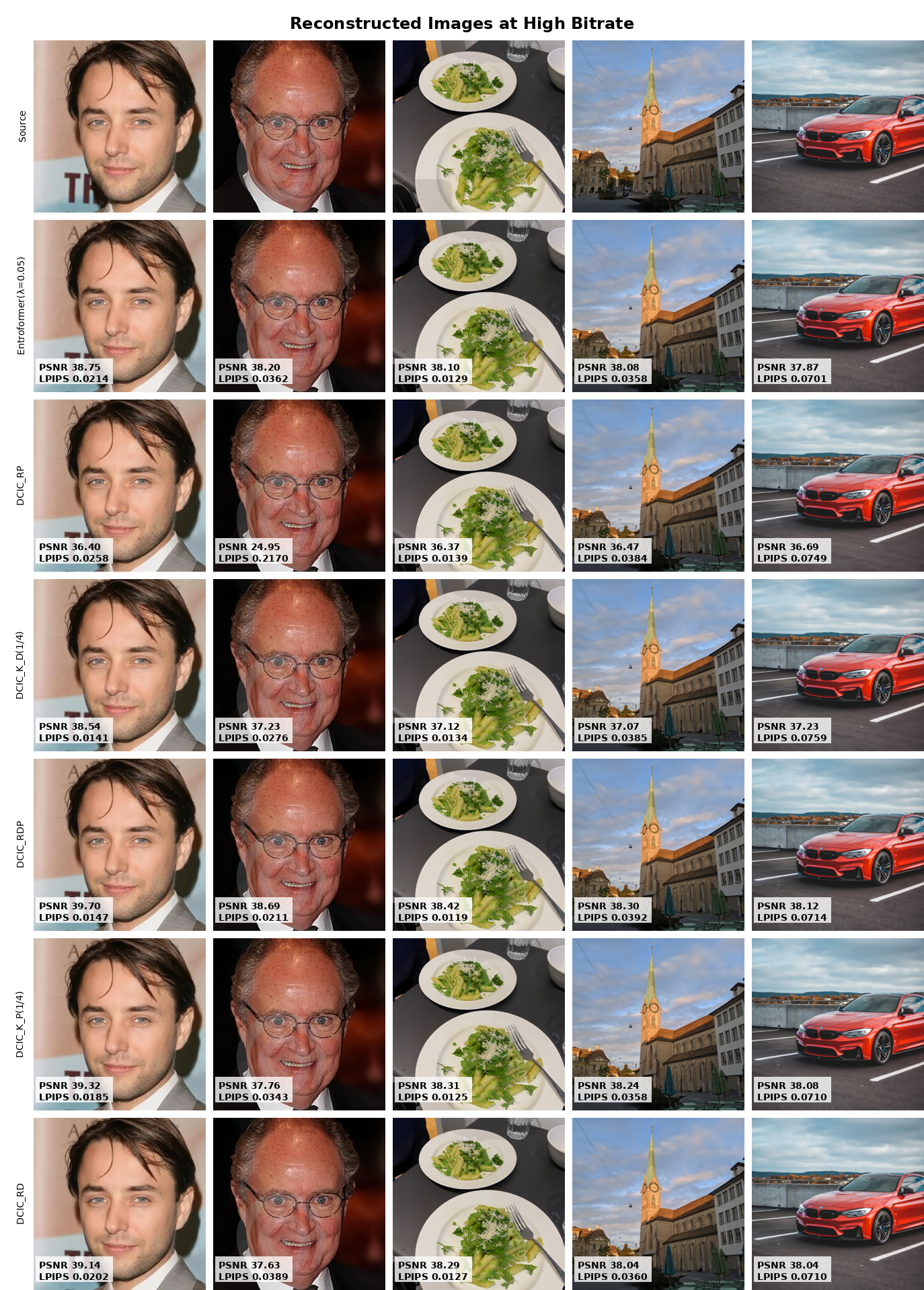}

\caption{
	Visual comparison at high bitrate ($\lambda=0.05$). 
	Row 1: source images; Row 2: Entroformer reconstructions; 
	Rows 3--7: DCIC restorations under varying attenuation factors 
	($K_D$, $K_P$). The narrow quality range across DCIC variants reflects 
	the saturation of the distortion--perception trade-off at high bitrates.
}
\label{fig:demons01}
\end{figure}

\subsection{Individual Trade-off Analysis}
\label{sec:per_bitrate}


Fig.~\ref{fig:detailed_cureves} presents the RD and RP curves of 
$\text{DCIC}_\text{RDP}$, 
$\text{DCIC}_\text{RD}$, 
$\text{DCIC}_\text{RP}$, Entroformer, 
Hyperprior, and WebP on CLIC2020 (0--1.0~bpp); 
PIC methods are excluded as they target 0.15--0.45~bpp and 
would be visually negligible across the full range. 

It can be observed that, 
$\text{DCIC}_\text{RDP}$ consistently achieves the highest fidelity across all bitrates, with the advantage widening at higher rates — on CLIC2020, the PSNR gain over Entroformer grows from $\sim$0.1~dB at 0.1~bpp to $\sim$0.6~dB at 1.0~bpp. This is because higher bitrates yield more precise codec representations, producing gradient estimates closer to unity and enabling the distortion constraint to approach its optimal extreme more accurately. Conversely, $\text{DCIC}_\text{RP}$
achieves the best realism at all bitrates, most markedly at low bitrates — 
consistent with the low-bitrate emphasis of PIC methods. 
As bitrate increases, conventional codecs exhibit rapidly improving FID, 
eventually approaching 
$\text{DCIC}_\text{RP}$, reflecting the shared idempotent nature of both approaches.

\subsection{Iterative Optimization Dynamics}
\label{sec:dynamics}

\begin{figure}[!htb]
\centering
\includegraphics[width=\linewidth]{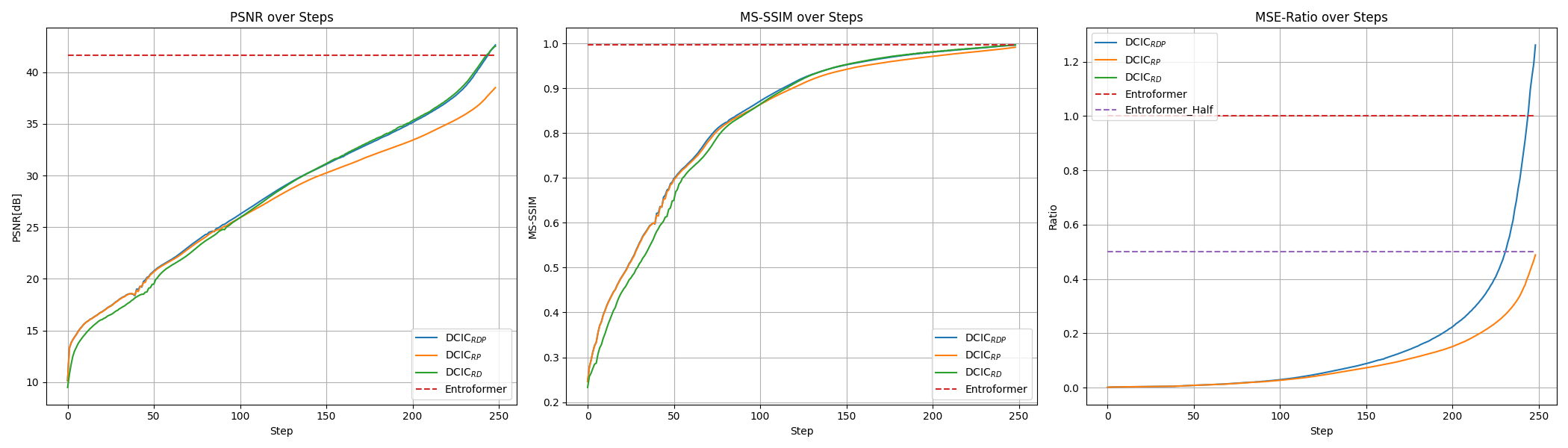}
\caption{
	Evolution of distortion metrics during the reverse denoising process of DCIC (Entroformer, $\lambda=0.02$). $\text{DCIC}_\text{RDP}$, $\text{DCIC}_\text{RD}$, and $\text{DCIC}_\text{RP}$ correspond to joint constraints $\mathcal{C}_\text{D}\cap\mathcal{C}_\text{P}$, distortion-only $\mathcal{C}_\text{D}$, and perception-only $\mathcal{C}_\text{P}$, respectively. (a) PSNR; (b) MS-SSIM; (c) MSE-Ratio. Dashed line: Entroformer baseline.
}
\label{fig:dynamics}
\end{figure}

Fig.~\ref{fig:dynamics} evaluates the impact of $\mathcal{C}_\text{D}$, $\mathcal{C}_\text{P}$, 
and their combination on the reverse denoising process, plotting PSNR, MS-SSIM, and MSE-Ratio — 
defined as the MSE of the DCIC-restored image relative to that of the Entroformer reconstruction — 
as functions of the diffusion time step for $\text{DCIC}_\text{RDP}$, $\text{DCIC}_\text{RD}$, and $\text{DCIC}_\text{RP}$.

All three metrics improve monotonically as the time step decreases, peaking at step zero. 
$\text{DCIC}_\text{RDP}$ and $\text{DCIC}_\text{RD}$ 
consistently meet or exceed the Entroformer baseline across all metrics, 
whereas $\text{DCIC}_\text{RP}$ exhibits markedly slower improvement below step $\sim$150 — 
confirming that the distortion constraint $\mathcal{C}_\text{D}$
is essential for maintaining fidelity throughout the reverse denoising process.

\subsection{Generalizability Study}
\label{sec:generalizability}

To evaluate generalizability, we instantiate DCIC with five additional LIC base codecs spanning CNN, Transformer, and hybrid CNN--Transformer architectures, evaluating all three configurations ($\text{DCIC}_\text{RDP}$, 
$\text{DCIC}_\text{RD}$, 
$\text{DCIC}_\text{RP}$) on CelebA-HQ (Table~\ref{tab:generalize}). 
Across all architectures, DCIC$_{\text{RDP}}$ consistently achieves 
the highest PSNR and MS-SSIM, $\text{DCIC}_\text{RP}$ the lowest LPIPS and FID, 
and $\text{DCIC}_\text{RD}$ performance closely aligned with the respective base codecs — 
confirming that DCIC realizes the RDP trade-off across diverse LIC architectures. 
The sole architectural requirement is continuous differentiability of the base codec function; 
standard non-differentiable codecs are therefore incompatible with the DCIC framework.

\begin{table*}[!th]
	\begin{center}
		\caption{
			Performance of DCIC instantiated with different LIC codecs, covering three representative architectures: CNN, Transformer, and Hybrid CNN–Transformer. 
		}
		\label{tab:generalize}
		\small
		\setlength{\tabcolsep}{3.5pt}  
		\begin{tabular}{l l  l  c  c  c  c c}
			\toprule
			\multicolumn{3}{c}{Image Compression Methods} 
			& \multicolumn{5}{c}{Performance Index} \\
			\hline
			Framework & Base Codec & DCIC 
			& PSNR$\uparrow$ &   MS-SSIM$\uparrow$ & LPIPS$\downarrow$   & FID$\downarrow$ & bpp\\
			\midrule 
			\multirow{11}{*}{CNN}&
			\multirow{4}{*}{Conv-Hyper.\cite{zhu2022transformer}} 
			&Base & 34.81 & $\bold{0.9828}$ & 0.0797     & 35.46 
			& \multirow{4}{*}{0.2576 } \\
			& &$\text{DCIC}_\text{RP}$ &32.23 & 0.9706 & $\bold{0.047}$  & $\bold{8.52}$  &\\ 
			& &$\text{DCIC}_\text{RD}$ &34.82 & 0.9823 & 0.0833   & 31.55 &\\ 
			&  &$\text{DCIC}_\text{RDP}$ &$\bold{34.86}$ & ${0.9827}$  & 0.0815  & 30.49  &\\  
			\cline{2-8}
			& \multirow{4}{*}{Conv-ChARM\cite{zhu2022transformer}} 
			&Base & 34.97 & 0.9828 & 0.0813   & 42.04 
			& \multirow{4}{*}{0.2443 } \\
			& &$\text{DCIC}_\text{RP}$ &32.56 & 0.97214 & $\bold{0.0430}$ & $\bold{8.83}$ &\\ 
			& &$\text{DCIC}_\text{RD}$ &34.99 & 0.9824 & 0.0847 & 34.17  &\\ 
			&  &$\text{DCIC}_\text{RDP}$ &$\bold{35.04}$ & $\bold{0.9829}$  & 0.0831 &  33.97  &\\ %
			\hline 
			\multirow{8}{*}{Trans.}&\multirow{4}{*}{SwinT-Hyper.\cite{zhu2022transformer}}
			&Base & 34.65 & $\bold{0.9826}$ & 0.0827    & 38.02
			& \multirow{4}{*}{0.2363 } \\
			&  &$\text{DCIC}_\text{RP}$ &31.80 & 0.9698 & $\bold{0.0519}$   & $\bold{9.48}$   &\\ 
			&  &$\text{DCIC}_\text{RD}$ &34.59 & 0.9819 & 0.0834   & 32.77   &\\ 
			&  &$\text{DCIC}_\text{RDP}$ &$\bold{34.66}$ & ${0.9824}$  & 0.0816   &  32.09   &\\  
			\cline{2-8}
			& \multirow{4}{*}{SwinT-ChARM\cite{zhu2022transformer}} 
			&Base & 34.91 & 0.9825 & 0.0834    & 42.15
			& \multirow{4}{*}{0.2221 } \\
			& &$\text{DCIC}_\text{RP}$ &32.70 & 0.9726 & $\bold{0.0436}$ & $\bold{9.13}$   & \\ 
			& &$\text{DCIC}_\text{RD}$ &34.88 & 0.9824 & 0.0821   & 34.53   &\\ 
			&   &$\text{DCIC}_\text{RDP}$ &$\bold{34.93}$ & $\bold{0.9826}$  & 0.0807 & 34.47  &\\ 
			\hline
			\multirow{4}{*}{Hybrid} & \multirow{4}{*}{TCM\cite{wu2020hybrid}} 
			&Base & 35.83 & $\bold{0.9856}$ & 0.0652 & 34.65 
			& \multirow{4}{*}{0.2530 } \\
			&  &$\text{DCIC}_\text{RP}$ &33.19 & 0.9747 & $\bold{0.0413}$  & $\bold{8.89}$    &\\ 
			&  &$\text{DCIC}_\text{RD}$ &35.80 & 0.9788  & 0.0736  & 30.81    &\\ 
			&  &$\text{DCIC}_\text{RDP}$ &$\bold{35.89}$ & ${0.9855}$  & 0.0664 & 29.37   & \\ 
			\bottomrule
		\end{tabular}
	\end{center}
\end{table*}

\subsection{Failure Case Study}

\begin{figure}[!tb]
	\centering
	\includegraphics[width=\linewidth]{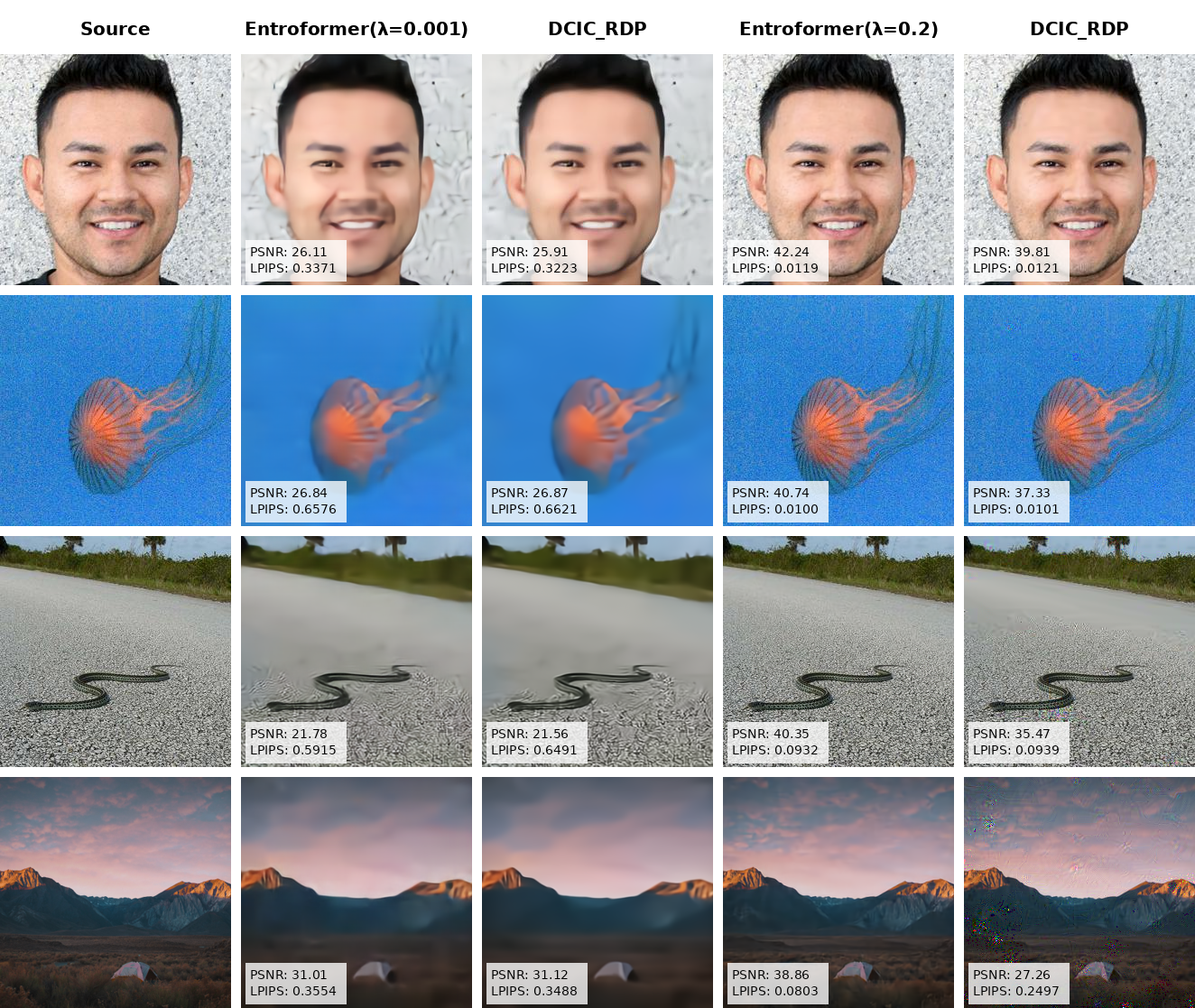}
	\caption{
		Typical failed samples of DCIC$_{\text{RDP}}$ and their corresponding reconstructions,  
		generated by the Entroformer,  
		covering three test sets: CelebA-HQ (first row), CLIC2020 (second row), and ImageNet-1K (the last two rows).}
	\label{fig:abnormals}
\end{figure}

Fig.~\ref{fig:abnormals} presents representative failure cases of $\text{DCIC}_\text{RDP}$, 
which predominantly involve substantial background noise. 
This limitation arises because the iterative optimization of $\text{DCIC}_\text{RDP}$ 
functions as a denoising procedure: while the mean and variance of noise can be reliably estimated, 
recovering large-scale granular noise — as seen in rows 1, 2, and 3 — remains inherently challenging.

$\text{DCIC}_\text{RDP}$ also exhibits greater robustness to noise at low bitrates than at high bitrates. %
At high bitrates, the guidance reconstructions (Entroformer outputs, column 4) 
reproduce background noise with greater clarity and finer detail, complicating denoising; 
at low bitrates, the blurrier reconstructions (column 2) make noise suppression comparatively easier.
Additionally, excessively large gradient steps during iterative optimization can induce 
oscillations or complete reconstruction failure, manifesting as prominent block artifacts — 
a risk that increases with bitrate, as exemplified by row 4 of Fig.~\ref{fig:abnormals}.

\section{Full Algorithm Pseudocode}
\label{sec:algorithm}

Algorithm~\ref{alg:roic} provides the complete DCIC decoding process
with line-by-line annotations.

\begin{algorithm}[t]
\caption{DCIC Decoding Process}
\label{alg:roic}
\begin{algorithmic}[1]
	\Require Bitstream $y$ (from base encoder $g_a$);
	attenuation factors $K_D,K_P\in[0,1]$;
	pretrained codec $\gc$, denoiser $\epsth$
	\Ensure  Restored image $\tilde{x}_0$
	\State $\bxhat \leftarrow g_s(y)$ \Comment{Base codec decode}
	\State $x_T \leftarrow \text{sample from }\mathcal{N}(0,I)$
	\Comment{Initialise with Gaussian noise}
	\State Precompute $\eta(t),\,\lambda^{\mathrm{OPT}}_D(t),\,
	\lambda^{\mathrm{OPT}}_P(t),\,\lambda_M(t)$ for
	$t=T,\ldots,1$ 
	\For{$t = T$ \textbf{down to} $1$}
	\State $\eta \leftarrow \eta(t);\;
	\lambda_D \leftarrow K_D\cdot\lambda^{\mathrm{OPT}}_D(t);\;
	\lambda_P \leftarrow K_P\cdot\lambda^{\mathrm{OPT}}_P(t);\;
	\lambda_M \leftarrow \lambda_M(t)$
	\State $\hat\epsilon_\theta \leftarrow \epsth(x_t,t)$
	\Comment{Predict noise}
	\State $\tilde{x}_0 \leftarrow
	(x_t - \sqrt{1-\bar\alpha_t}\,\hat\epsilon_\theta)/\sqrt{\bar\alpha_t}$
	\Comment{DDIM one-step clean estimate}
	\State $J^{(t)} \leftarrow
	\lambda_D\|\bxhat-\tilde{x}_0\|^2
	+\lambda_P\|\bxhat-\gc(\tilde{x}_0)\|^2
	+\lambda_M\|x_t-\tilde\mu_t\|^2$
	\State $x_t \leftarrow x_t
	+ \eta\cdot\nabla_{x_t}J^{(t)}$
	\Comment{Gradient update} 
	\State $\hat\epsilon_\theta \leftarrow \epsth(x_t,t)$
	\Comment{Re-predict after update}
	\State $\tilde{x}_0 \leftarrow
	(x_t - \sqrt{1-\bar\alpha_t}\,\hat\epsilon_\theta)/\sqrt{\bar\alpha_t}$
	\State $x_{t-1} \leftarrow \text{DDIM\_step}(x_t,\tilde{x}_0,t)$
	\Comment{DDIM reverse sampling step}
	\EndFor
	\State \Return $\tilde{x}_0$
\end{algorithmic}
\end{algorithm}

\noindent\textbf{Implementation notes.}
\begin{enumerate}
\item Line~9 performs gradient ascent on the log-probability
(equivalently, gradient descent on $J^{(t)}$), implemented via
automatic differentiation through $\gc$.
\item The base codec $\gc$ must be continuously differentiable;
standard non-differentiable codecs (e.g., JPEG) cannot be used.
\item The double forward pass (lines~6--7 then 10--11) is necessary
because line~9 modifies $x_t$, requiring updated estimates before
the DDIM step.
\item Special cases recover standard variants:
setting $K_P=0$ gives $\ROICRD$;
setting $K_D=0$ gives $\ROICRP$;
both equal to 1 gives $\ROICRDP$.
\end{enumerate}

\section{Additional Ablation Studies}
\label{sec:ablations}

\subsection{Effect of Number of Diffusion Steps $T$}
\label{sec:ablation_T}

Table~\ref{tab:T} evaluates $\ROICRDP$ across $T\in\{50, 100,200,250,300, 500,800\}$
on CLIC2020 (Entroformer, $\lambda=0.01$). Fidelity peaks between $T=200$ and $T=300$, 
while perceptual quality (LPIPS) improves monotonically with $T$ at the cost of proportionally increasing inference time. 
Balancing restoration quality against computational overhead, we set $T=250$.


\begin{table}[t]
\centering
\caption{Effect of number of diffusion steps $T$ on $\ROICRDP$
	(CLIC2020, $\lambda\!=\!0.01$).} 
	\label{tab:T}
	\setlength{\tabcolsep}{10pt}
	\small
	\begin{tabular}{lcccc}
\toprule
$T$ & PSNR (dB)$\uparrow$ & MS-SSIM$\uparrow$ & LPIPS$\downarrow$ & Time (s)$\downarrow$ \\
\midrule
50    & 32.32 & 0.9702 & 0.1323 & $\approx$13 \\
100  & 33.92 & 0.9782 & 0.1149 & $\approx$27 \\
200  & 34.24 & 0.9810 & 0.1109 & $\approx$48 \\
250  & \textbf{34.26} & {0.9812} & {0.1110} & $\approx$63 \\ %
300  & 34.18 & \textbf{0.9813} & 0.1108 & $\approx$87 \\
500  & 34.05 & 0.9812 & 0.1095 & $\approx$119 \\
800  & 33.98 & 0.9810 & \textbf{0.1083} & $\approx$218 \\
\bottomrule
\end{tabular}
\end{table}

\subsection{Effect of Learning Rate Schedule Shape}
\label{sec:ablation_lr}

Table~\ref{tab:lr} compares three learning rate schedule shapes for
$\eta(t)$: constant, linear ramp, and our gamma-distribution segment.
The gamma schedule provides the best PSNR/FID balance by combining
small learning rates in the high-noise regime (preventing instability
from large $1/\sqrt{\bar\alpha_t}$ amplification) with larger rates
in the low-noise regime (accelerating convergence).

\begin{table}[t]
\centering
\caption{Comparison of learning rate schedules
($\ROICRDP$, CLIC2020, $T\!=\!250$, bpp~$\approx\!0.25$).}
\label{tab:lr}
\setlength{\tabcolsep}{12pt}
\small
\begin{tabular}{lccc}
\toprule
Schedule & PSNR (dB)$\uparrow$ & MS-SSIM$\uparrow$ &LPIPS$\downarrow$ \\
\midrule
Constant    & 32.83 & 0.9677 & 0.1336\\
Linear ramp       & 33.61 & 0.9728 & 0.1164 \\
Gamma (ours)      & \textbf{33.76} & \textbf{0.9745} & \textbf{0.1065}\\
\bottomrule
\end{tabular}
\end{table}

%

\subsection{Sensitivity to Weighting Function Shape ($\sigma$)}
\label{sec:ablation_sigma}

Table~\ref{tab:ws} ablates $\sigma\in\{1.5, 2.5, 3.5, 4.5, 5.5\}$ for the weighting function $\lambda(x)=(k/\sigma\sqrt{2\pi})\exp(-x^2/2\sigma^2)$ on CLIC2020 ($\ROICRDP$). 
The optimal value $\sigma=3.5$, used in all main experiments, 
provides the best fidelity--perception balance: smaller $\sigma$ concentrates weight too early, 
providing insufficient late-stage guidance, while larger $\sigma$ flattens the weighting curve toward a constant, diminishing its adaptive effect and weakening the perceptual constraint as $t\to 0$.


\begin{table}[htb]
\centering
\caption{Comparison of weighting function shapes
($\ROICRDP$, CLIC2020, $T\!=\!250$, bpp~$\approx\!0.24$).}
\label{tab:ws}
\setlength{\tabcolsep}{12pt}
\small
\begin{tabular}{lccc}
\toprule
Weighting Shape($\sigma$) & PSNR (dB)$\uparrow$ & MS-SSIM$\uparrow$ &LPIPS$\downarrow$ \\
\midrule
1.5    & 33.733 & 0.9914 & \textbf{0.0273}\\
2.5       & 33.742 & 0.9914 & 0.0281 \\
3.5 (ours)      & \textbf{33.745} & \textbf{0.9914} & {0.0288}\\
4.5    & 33.743 & 0.9913 & 0.0293\\
5.5       & 33.734 & 0.9913 & 0.0297 \\
\bottomrule
\end{tabular}
\end{table}

\section{Implementation Details and Reproducibility}
\label{sec:impl}

\subsection{Base Codec Configurations}
\label{sec:impl_codec}

All base codecs use official pretrained weights. For Entroformer, 
quality parameters $\lambda\in\{0.001,0.002,0.005,0.01,0.02,0.05\}$ correspond to 
bitrates $\approx$0.12--0.98~bpp on CLIC2020. Base codec reconstructions $\hat{\mathbf{x}}$
are computed once and cached prior to DCIC decoding to avoid redundant forward passes.


\subsection{Diffusion Model Configuration}
\label{sec:impl_diff}

\begin{itemize}
\item \textbf{CelebA-HQ:} 256$\times$256 face diffusion model of
Lugmayr et al.~\cite{lugmayr2022repaint} (U-Net backbone,
classifier-free guidance).
\item \textbf{CLIC2020 and ImageNet-1K:} ADM model of Dhariwal and
Nichol~\cite{dhariwal2021diffusion} trained on ImageNet
256$\times$256, with classifier guidance scale~$=2.5$.
\end{itemize}

All diffusion model parameters are frozen in \texttt{eval()} mode;
gradients are computed only w.r.t.\ the noisy sample $x_t$.

\subsection{Hyperparameter Settings}
\label{sec:impl_hyper}

Table~\ref{tab:hyper} lists all hyperparameters.
Validation-set tuning proceeds by fixing one constraint
($\cD$ or $\cP$) and optimizing the other.

\begin{table}[t]
\centering
\caption{Hyperparameter settings for DCIC on each dataset ($\lambda=0.01$).}
\label{tab:hyper}
\setlength{\tabcolsep}{6pt}
\small
\begin{tabular}{lccc}
\toprule
Hyperparameter & CelebA-HQ & CLIC2020 & ImageNet-1K \\
\midrule
$T$ (reverse steps)              & 250  & 250  & 250  \\
$\eta$ schedule: $k$             & 2.65 & 2.55 & 2.55 \\
$\eta$ schedule: $\theta$        & 1.85 & 1.50 & 1.50 \\
$\lambda_D$ schedule: $\sigma$   & 3.5  & 3.5  & 3.5  \\
$\lambda_D$ schedule: $k$        & 0.32    & 0.3    & 0.37    \\
$\lambda_P$ schedule: $\sigma$   & 3.5  & 3.5  & 3.5  \\
$\lambda_P$ schedule: $k$        & 3.8  & 2.2  & 1.8  \\
$K_D$ ($\ROICRDP$)               & 1.0  & 1.0  & 1.0  \\
$K_P$ ($\ROICRDP$)               & 1.0  & 1.0  & 1.0  \\
Validation images                & 50   & 50   & 50   \\
Test images (evaluation)         & 500  & 500  & 500  \\
\bottomrule
\end{tabular}
\end{table}

\subsection{Hardware and Runtime}
\label{sec:impl_hw}

All experiments are conducted on a single NVIDIA A100 80\,GB GPU.
Base codec encoding/decoding uses CPU for compatibility with
non-differentiable quantization in standard LIC implementations;
gradient computation through $\gc$ is performed on GPU using a
differentiable surrogate forward pass.
Average runtime per 256$\times$256 image: $\sim$63\,s (DCIC)
vs.\ $\sim$0.2\,s (base codec alone).
GPU VRAM usage: $\sim$18\,GB (ImageNet ADM model + gradient buffers).

\subsection{Evaluation Protocol Clarifications}
\label{sec:impl_eval}

\begin{itemize}
\item All images are center-cropped to 256$\times$256 prior to
compression and evaluation.
\item FID is computed between the full test set of
restored outputs and the corresponding source images, using
Inception-v3 features. 
\item LPIPS uses the AlexNet backbone (Zhang et al.~\cite{zhang2018lpips})
in full-reference mode.
\item BD metrics use the Bj\o ntegaard delta method~\cite{bjontegaard2001psnr}
across six rate points ($\lambda\in\{0.001,0.002,0.005,0.01,0.02,0.05\}$).
\item For non-DCIC baselines, official pretrained weights and
published evaluation protocols are used.
\end{itemize}

\section{Extended Limitations and Future Directions}
\label{sec:limitations}

\paragraph{Inference latency.}
The 250-step reverse diffusion process requires $\sim$63\,s per
256$\times$256 image, which is impractical for real-time decoding.
Accelerated samplers such as DEIS\cite{zhang2022fast}, DPM-Solver\cite{lu2022dpm}\cite{lu2025dpm}, or consistency
distillation could reduce this to $\sim$5--10 steps ($\sim$25$\times$
speedup) with minimal quality loss.

\paragraph{Low-bitrate gradient approximation.}
The approximation $\partial\gc/\partial\tilde{x}_0\approx 1$ degrades
at very low bitrates ($\lambda<0.005$), where the heavily quantized
codec output poorly tracks input variations.
Bitrate-conditioned gradient scaling or second-order correction terms
could improve DCIC in the sub-0.15~bpp regime.

\paragraph{Semantic-dependent convergence.}
DCIC performs best on CelebA-HQ (faces) and worst on ImageNet-1K
(diverse categories), consistent with the theoretical
rate--distortion--perception--semantics
trade-off~\cite{wang2024semantics}.
Semantic-aware weight functions $\lambda_D(t),\lambda_P(t)$ conditioned
on features extracted from $\bxhat$ (e.g., via CLIP embeddings) could
close this gap by adapting the optimization trajectory to content
complexity.

\paragraph{Extension to video compression.}
The DCIC framework naturally extends to video by conditioning the
diffusion model on both the spatial codec reconstruction and temporal
motion information.
The idempotence constraint generalizes to temporal idempotence
(re-encoding a restored frame recovers the original compressed frame),
providing a principled foundation for RDP-optimal video compression.

\section{Broader impacts}
\label{sec:broader_impacts}

Positive impacts include improved bandwidth
efficiency and perceptual quality for image transmission in
bandwidth-constrained environments (e.g., medical imaging,
remote sensing, and video streaming). Potential negative impacts
are also acknowledged: perceptually realistic reconstructions
could lower the barrier to producing visually plausible but
semantically altered images, with implications for
misinformation.

\end{document}